\documentclass[journal,final,twoside]{IEEEtran}
\usepackage{cite}

\ifCLASSINFOpdf
   \usepackage[pdftex]{graphicx}
   \graphicspath{{./figures/}{./pdf/}{./jpeg/}}
   \DeclareGraphicsExtensions{.pdf,.jpeg,.png}
\else
   \usepackage[dvips]{graphicx}
   \graphicspath{{./eps/}}
   \DeclareGraphicsExtensions{.eps}
\fi
\usepackage{algorithm}
\usepackage{algpseudocode}

\usepackage[cmex10]{amsmath}
\usepackage{amssymb}

\usepackage[caption=false,font=footnotesize]{subfig}
\usepackage{stfloats}

\usepackage{multirow}

\usepackage{footnote}

\usepackage{threeparttable}

\renewcommand{\vec}[1]{\mathbf{#1}}

\newcommand{\mtx}[1]{\mathbf{#1}}

\newcommand{\transU}[1]{\hat{#1}}

\begin{document}
%
\title{Postprocessing of Compressed Images via\\Sequential Denoising}
%
%
%

\author{Yehuda Dar, Alfred M. Bruckstein, Michael Elad, and Raja Giryes
\thanks{EDICS: Restoration and Enhancement (TEC-RST), Lossy Coding of Images and Video (COM-LOC).}
\\
\thanks{Y. Dar, A. M. Bruckstein, and M. Elad are with the Department of Computer Science, Technion, Israel. E-mail addresses: \{ydar,~freddy,~elad\}@cs.technion.ac.il. R. Giryes is with the Department of Electrical and Computer Engineering, Duke University, USA. E-mail address: raja.giryes@duke.edu.}
\thanks{The research leading to these results has received funding from the European Research Council under European Union’s Seventh Framework Program, ERC Grant agreement no. 320649.}
}

%
%

\markboth{}%
{Postprocessing of Compressed Images via Sequential Denoising}
%



\maketitle

\begin{abstract}

In this work we propose a novel postprocessing technique for compression-artifact reduction. Our approach is based on posing this task as an inverse problem, with a regularization that leverages on existing state-of-the-art image denoising algorithms. We rely on the recently proposed Plug-and-Play Prior framework, suggesting the solution of general inverse problems via Alternating  Direction Method of Multipliers (ADMM), leading to a sequence of Gaussian denoising steps. A key feature in our scheme is a linearization of the compression-decompression process, so as to get a formulation that can be optimized. In addition, we supply a thorough analysis of this linear approximation for several basic compression procedures.
The proposed method is suitable for diverse compression techniques that rely on transform coding. Specifically, we demonstrate impressive gains in image quality for several leading compression methods - JPEG, JPEG2000, and HEVC. 
\end{abstract}



%
\IEEEpeerreviewmaketitle

\section{Introduction}
\IEEEPARstart{B}{andwidth} and memory constraints play a crucial role in transmission and storage systems.
Various compression methods are available in order to meet severe constraints on the bit-cost in data representation. While some applications require perfect reconstruction, some may tolerate inaccuracies and can benefit from a reduced representation-cost.
The latter approach is known as lossy compression and is widely used for representing a signal under bit-budget constraints while allowing some errors in recovery. Accordingly, a variety of techniques were standardized over the years for the lossy compression of acoustic and visual signals.

Since lossy compression allows discrepancies between the original and the reconstructed signals, the differences being intentionally used in tradeoffs between bit-rate and quality. The nature of the created artifacts depends on the compression architecture. For example, block-based image compression techniques suffer from blockiness effects that increase and degrade the reconstruction as the bit-rate is reduced.

As artifacts are inherent in the lossy compression of signals, a great number of artifact-reduction techniques were proposed over the years (e.g., \cite{RefWorks:141,RefWorks:143,RefWorks:144,RefWorks:146,RefWorks:147,RefWorks:148,RefWorks:150,RefWorks:151,RefWorks:152,RefWorks:153,RefWorks:154,RefWorks:155,RefWorks:156,RefWorks:157,RefWorks:159,RefWorks:160,RefWorks:149,foi2007pointwise,zhang2013compression,zhai2009efficient,Kwon2015Efficient} for image compression). These methods usually focus on specific signal types (e.g., image, video or audio) and sometimes even on specific artifacts corresponding to certain compression designs (e.g., deblocking procedures for images).
Common image compression techniques rely on transform-coding, where image blocks are transformed, and the resultant transform-coefficients are quantized according to their relative importance.
The prominent artifacts of this architecture are \cite{RefWorks:138}: blockiness due to the separate treatment of non-overlapping blocks; ringing caused by the effective elimination of high frequency components, expressed as contours spreading along sharp edges; and blurring that results from high-frequency information loss.
Postprocessing of compressed images are subcategorized into two approaches \cite{RefWorks:138}: enhancement of the deteriorated signal by smoothing its artifacts (e.g., \cite{RefWorks:144,RefWorks:141}), and restoration of the original signal samples (e.g., \cite{RefWorks:159,RefWorks:148}).

In this work we propose a novel postprocessing technique for compression artifact reduction by a regularized restoration of the original (precompressed) signal.
Specifically, we formulate the compression postprocessing procedure as a regularized inverse-problem for estimating the original signal given its reconstructed form.
We also approximate the (nonlinear!) compression-decompression process by a linear operator, so as to obtain a tractable inverse problem formulation.
The intriguing approach of locally linearizing the non-differentiable compression procedures is carefully analyzed, in order to utilize it properly.
Whereas many studies focus on corrections of specific artifacts (e.g., image deblocking techniques \cite{RefWorks:144,RefWorks:141,RefWorks:156,RefWorks:147}), our approach attempts to generally restore the signal and thus implicitly repairs multiple artifacts. 
The major strength of our method comes from the regularization used, as we next explain.

Afonso et al. \cite{RefWorks:167} proposed to efficiently solve regularized inverse-problems in image processing using the Alternating Direction Method of Multipliers (ADMM) \cite{RefWorks:166}. Their approach decouples the inversion and the regularization parts of the optimization problem, which is in turn iteratively solved. 
Venkatakrishnan et al. \cite{RefWorks:168} further developed the use of the ADMM by showing an equivalence between the regularization step and denoising optimization problems. Their framework, called "Plug-and-Play Priors", is flexible, proposing the replacement of the regularization step by a general-purpose Gaussian image denoiser.

In this work we propose a compression postprocessing algorithm by employing the Plug-and-Play Priors framework. 
Furthermore, as denoising algorithms relying on sparse models were found to be highly effective ones (e.g., K-SVD \cite{aharon2006KSVD,RefWorks:170}, BM3D \cite{RefWorks:169}), we utilize a leading denoiser from this category.
The Plug-and-Play Priors framework was proposed for general inverse problems and was specifically demonstrated for reconstruction of tomographic images.
The novelty of our work with respect to the original Plug-and-Play approach is that we apply it for the task of compression-artifact reduction. Moreover, we utilize it to address an inverse-problem for a forward-model that is non-linear and non-differentiable.

Since we propose a method of postprocessing for a variety of lossy-compression techniques, the algorithm and its analysis are dealt within an abstract and general setting. Following that, a thorough demonstration for image compression is provided.
Specifically, we show results for the leading image compression standards: JPEG \cite{RefWorks:162}, JPEG2000 \cite{RefWorks:163} and the still-image profile of the HEVC \cite{RefWorks:112,RefWorks:164}, offering the state-of-the-art performance \cite{RefWorks:165}.
While these three compression methods rely on a block-based architecture and a transform-coding approach, they differ as follows: JPEG operates on 8x8 blocks and applies a discrete cosine transform (DCT); JPEG2000 works on large blocks (tiles) of at least 128x128 pixels and utilizes a discrete wavelet transform (DWT); in HEVC-stills the image is split into coding blocks that are further partitioned using a quadtree structure, then intra-prediction is performed and transform coding is applied on the prediction residuals (where the transform is mainly integer-approximations of the DCT at various sizes).
Our method is evaluated for a diversified set of compression algorithms that span the range of the contemporary coding concepts. Moreover, our postprocessing technique achieves significant gains and usually outperforms the cutting-edge methods for the examined compression standards.

This paper is organized as follows. In section II the proposed postprocessing method is presented. In section III, the compression linearization is mathematically analyzed for simplified cases of quantization and transform coding. Section IV presents image-compression experimental results and compares them to those of competitive techniques. Section V concludes this paper.

\section{The Proposed Postprocessing Strategy}
\label{sec:The Proposed Postprocessing Strategy}
\subsection{Problem Formulation using ADMM}
Let us consider a signal $ \vec{x} \in \mathbb{R} ^N $ that undergoes a compression-decompression procedure, $ C:{\mathbb{R}^N} \to {\mathbb{R}^N} $, resulting in the reconstructed signal $ \vec{y} = C\left( \vec{x} \right) $. For lossy compression methods an error is introduced at a size that depends on the bit-budget, the specific-signal characteristics, and the compression algorithm. We aim at restoring the precompressed signal $ \vec{x} $ from the reconstructed $ \vec{y} $ using the following regularized inverse-problem:
\begin{IEEEeqnarray}{rCl}
	\label{eq:basic regularized inverse problem}
	\hat{\vec{x}} = \mathop {\arg \min }\limits_{\vec{x}} \,\,\left\| {\vec{y} - C\left( \vec{x} \right)} \right\|_2^2 + \beta s\left( \vec{x} \right) ,
\end{IEEEeqnarray}
where $ s\left( \cdot{} \right) $ is a regularizer, which can be associated with a given Gaussian denoiser, weighted by the parameter $\beta$.
For example, assuming that the image is piecewise constant promotes the utilization of the popular total-variation regularizer, $s(\vec{x}) = || \vec{x} ||_{TV}$ \cite{rudin1992nonlinear}.

One should note that $\vec{y}$ and $C\left( \vec{x} \right)$ are two signals reconstructed from compression, and therefore, the fidelity term in Equation (\ref{eq:basic regularized inverse problem}) expresses their distance. Notice that this is substantially different from $\left\| {\vec{y} - \vec{x} } \right\|_2^2$ -- whereas the latter has compression artifacts as error, the one we deploy represents a milder distortion. Throughout this paper we shall assume for simplicity that the distortion between the two reconstructions, $\vec{y}$ and $C\left( \vec{x} \right)$, is modeled as a white additive Gaussian noise, leading to the $\ell_2$ term used here. We should note, however, that our scheme could be improved by using a better modeling of the reconstructed-signal error, such as an $\ell_\infty$ on the transform coefficients w.r.t. the quantization step-size (in the case of transform coding).

Similar to \cite{RefWorks:167} and \cite{RefWorks:168}, we develop an iterative algorithm for the solution of (\ref{eq:basic regularized inverse problem}).
We start by applying variable splitting that yields the following equivalent form of (\ref{eq:basic regularized inverse problem}):
\begin{IEEEeqnarray}{rCl}
	\label{eq:variable splitting form}
	&&\mathop {\min }\limits_{\vec{x},\vec{v}} \,\,\left\| {\vec{y} - C\left( \vec{x} \right)} \right\|_2^2 + \beta s\left( \vec{v} \right)
	\\ \nonumber
	&&\text{subject to   } \vec{x} = \vec{v} ,
\end{IEEEeqnarray}
where $\vec{v} \in \mathbb{R} ^N$ is an additional vector due to the split.
The constrained problem (\ref{eq:variable splitting form}) is addressed by forming an augmented Lagrangian and its corresponding iterative solution (of its scaled version) via the method of multipliers \cite[ch. 2]{RefWorks:166}, where the $i^{th}$ iteration consists of
\begin{IEEEeqnarray}{rCl}
	\label{eq:method of multipliers form}
	&& \left( {\hat{\vec{x}}}_i , \hat{ \vec{v} }_i  \right)  =  \mathop {\arg \min }\limits_{\vec{x},\vec{v}} \left\| {\vec{y} - C\left( \vec{x} \right)} \right\|_2^2 + \beta s\left( \vec{v} \right) 
	\\ \nonumber
	&&\qquad\qquad ~~~~~~~~~~~ + \frac{\lambda }{2} \left\| {\vec{x} - \vec{v} - {\vec{u}}_i} \right\|_2^2
	\\ \nonumber
	&& {\vec{u}}_{i+1} = {\vec{u}}_i + \left( {{\hat{\vec{x}}}_i - {\hat{\vec{v}}}_i} \right) .
\end{IEEEeqnarray}
Here ${\vec{u}}_i \in \mathbb{R} ^n$ is the scaled dual-variable and  $ \lambda $ is an auxiliary parameter, both introduced in the Lagrangian. 

Please note the following notation remark for a general vector $\vec{u}$. First, ${{\vec{u}}}_i$ stands for vector ${{\vec{u}}}$ in the $i^{th}$ iteration. On the other hand, $u_j$ represents the $j^{th}$ component (a scalar) of the vector $\vec{u}$. Finally, ${{\vec{u}}}_i^{(j)}$ denotes the $j^{th}$ element of the vector ${{\vec{u}}}_i$.

Approximating the joint optimization of $\vec{x}$ and $\vec{v}$ in (\ref{eq:method of multipliers form}), using one iteration of alternating minimization, results in the iterative solution in the ADMM form, where the $i^{th}$ iteration consists of
\begin{IEEEeqnarray}{rCl}
	\label{eq:iterative solution equations - forward model step}
	&& {\hat{\vec{x}}}_i  =  \mathop {\arg \min }\limits_{\vec{x}} \left\| {\vec{y} - C\left( {\vec{x}} \right)} \right\|_2^2 + \frac{\lambda }{2}\left\| { {\vec{x}} - \tilde{{\vec{x}}}_i} \right\|_2^2
	\\
	\label{eq:iterative solution equations - regularization step}
	&& {\hat{\vec{v}}}_i  =  \mathop {\arg \min} \limits_{\vec{v}} \frac{\lambda }{2}\left\| { {\vec{v}} - \tilde{\vec{v} }_i} \right\|_2^2 + \beta s\left( \vec{v} \right)
	\\
	\label{eq:iterative solution equations - updating u step}
	&& {\vec{u}}_{i+1} = {\vec{u}}_i + \left( {{\hat{\vec{x}}}_i - {\hat{\vec{v}}}_i} \right) .
\end{IEEEeqnarray}
Here $ \tilde{{\vec{x}}}_i = \hat{\vec{v}} _{i-1} - \vec{u}_i $ and $ \tilde{\vec{v} }_i = \hat{ \vec{x} }_{i} + \vec{u} _i $.

The regularization step (\ref{eq:iterative solution equations - regularization step}) is of the form of a Gaussian denoising optimization-problem (of a noise level determined by $ {\beta  \mathord{\left/ {\vphantom {\beta  \lambda }} \right.\kern-\nulldelimiterspace} \lambda } $) and therefore can be viewed as applying a denoising algorithm to the signal $\tilde{\vec{v} }_i$. 
More specifically, this corresponds to assuming that ${\tilde{\vec{v} }_i} = {{\vec{v} }} + \vec{w}$, where $\vec{w}$ is an i.i.d zero-mean Gaussian vector with variance $1/\lambda$ (and a corresponding distribution function denoted as $p_w(\vec{w})$). 
In addition, ${{\vec{v} }}$ is assumed to be drawn from a distribution $p_s({\vec{v} })$ that is proportional to $\exp(-\beta s(\vec{x}))$.
Then, the Maximum A-Posteriori (MAP) estimator of ${\vec{v} }$ from its (white Gaussian) noisy version $\tilde{\vec{v} }_i$ is formed as 
\begin{IEEEeqnarray}{rCl}
	\label{eq:MAP estimator}
	{\hat{\vec{v}}}_i  =  \mathop {\arg \max} \limits_{\vec{v}} \log{p_w({\tilde{\vec{v} }_i} - {{\vec{v} }})} + \log{p_s({\vec{v} })},
\end{IEEEeqnarray}
which for the above defined distribution functions, $p_w(\cdot)$ and $p_s(\cdot)$, is equivalent to (\ref{eq:iterative solution equations - regularization step}) and, thus, establishes the latter as a Gaussian denoising procedure.
Indeed, the Plug-and-Play Priors framework \cite{RefWorks:168} suggests exactly this strategy, replacing (\ref{eq:iterative solution equations - regularization step}) with an independent denoiser; 
even one that does not explicitly have in its formulation a minimization problem of the form of (\ref{eq:iterative solution equations - regularization step}).
The deployment of a favorable denoiser introduces valuable practical benefits to the design of the proposed postprocessing procedure, and yields a powerful generic method.

\subsection{Linear Approximation of the Compression-Decompression Procedure}

Due to the high nonlinearity of $ C\left( \vec{x} \right) $, we further simplify the forward-model step (\ref{eq:iterative solution equations - forward model step}) using a first-order Taylor approximation of the compression-decompression function around $ {{{\hat{\vec{x}}}_{i-1} }} $, i.e., 
\begin{IEEEeqnarray}{rCl}
	\label{eq:Linear approximation of the compression-decompression}
	{C_{lin}}\left( \vec{x} \right) = C\left( { {\hat{\vec{x}} }_{i-1}} \right) + \left.{ \frac{dC\left( \vec{z} \right)}{d\vec{z}} } \right|_{\vec{z}={\hat{\vec{x}}_{i-1}}} \cdot \left( {\vec{x} - {{\hat {\vec{x}}}_{i-1}} } \right)
\end{IEEEeqnarray}
where $\left.{\frac{dC\left( \vec{z} \right)}{d\vec{z}} }\right|_{\vec{z}={\hat{\vec{x}}_{i-1}}}$ is the ${N \times N}$ Jacobian matrix of the compression-decompression at the point $ {\hat{\vec{x}}}_{i-1} $. 

Since the approximation of the Jacobian, $ \frac{dC\left( \vec{z} \right)}{d\vec{z}} $, deeply influences the restoration result and the computational cost, this is a quite delicate task. First, $C$ is a non-linear and even non-differentiable function as the compression often relies on quantization and/or thresholding. Second, we provide here a generic technique, and therefore do not explicitly consider the compression-decompression formulation.

Theoretically, the fact that $C$ is non-differentiable would prevent us from using its Jacobian for the optimization. However, as its Jacobian has only a finite number of singularities, in practice we can rely on it as is done in other fields, e.g., in the training of neural networks that are composed of concatenations of non-differentiable non-linear operations \cite{Schmidhuber15Overview}. 

For calculating the entries of the Jacobian, we rely on the standard definition of the derivative, assuming that $C$ is locally linear.
We justify this approach in the next section. 
As we might be approximating the derivative in the neighborhood of a non-differential point, we take several step-sizes in the calculation of the derivative and average over all of them. This leads to the following approximation to the $k^{th}$ column of the Jacobian:
\begin{IEEEeqnarray}{rCl}
	\label{eq:Jacobian estimation - k-th column}
	\frac{dC\left( \vec{z} \right)}{dz_k} =\frac{1}{\left| S_{\delta} \right|} \sum\limits_{\delta \in S_{\delta}} {\frac{C( \vec{z}+ \delta \cdot \vec{e}_k ) - C(\vec{z} - \delta \cdot \vec{e}_k )}{2\delta}},
\end{IEEEeqnarray}
where $\vec{e}_k$ is the $k^{th}$ standard direction vector, and $S_{\delta}$ is a set of step lengths for approximating the derivative using the standard definition (the set size is denoted as $\left| S_{\delta} \right|$).

Due to the high nonlinearity of $C$, the linear approximation (\ref{eq:Linear approximation of the compression-decompression}) is reasonable in a small neighborhood around the approximating point $\hat{\vec{x}}_{i-1}$. Accordingly, we further constrain the distance of the solution from the linear-approximation point by modifying (\ref{eq:iterative solution equations - forward model step}) to
\begin{IEEEeqnarray}{rCl}
	\label{eq:iterative solution equations - forward model step - closeness constraint}
	\hat{\vec{x}}_i & = & \mathop {\arg \min }\limits_{\vec{x}} \left\| {\vec{y} - {C_{lin}} \left( \vec{x} \right)} \right\|_2^2 
	\\ \nonumber
	&& \qquad ~~~ + \frac{\lambda }{2}\left\| { \vec{x} - \tilde{\vec{x}}_i} \right\|_2^2 + \mu \left\| {\vec{x} - {\hat{\vec{x}}_{i-1}}} \right\|_2^2 .
\end{IEEEeqnarray}
The proposed generic method is summarized in Algorithm \ref{Proposed Postprocessing Method}.

\begin{algorithm}
	\caption{The Proposed Postprocessing Method}\label{Proposed Postprocessing Method}
	\begin{algorithmic}[1]
		\State $\hat {\vec{x}} _0 = \vec{y}$ ,  $\hat {\vec{v}} _0 = \vec{y}$
		\State $i = 1$, $\vec{u}_1 = \vec{0}$
\Repeat

\State Approximate $C_{lin}(\cdot)$ around ${\hat {\vec{x}} }_{i-1}$ using (\ref{eq:Linear approximation of the compression-decompression}) and (\ref{eq:Jacobian estimation - k-th column})
\State $ \tilde{\vec{x}} _i = \hat {\vec{v}} _{i-1} - \vec{u} _i $
		\State $\hat {\vec{x}} _i  = \mathop {\arg \min }\limits_{{\vec{x}}} \left\| {\vec{y} - {C_{lin}}\left( {\vec{x}} \right)} \right\|_2^2 +\nolinebreak \frac{\lambda }{2}\left\| {{\vec{x}} - \tilde {\vec{x}} _i} \right\|_2^2 \newline~~~~~~~~~~~~~~~~~~~~~ +\mu \left\| {{\vec{x}} - {\hat {\vec{x}}}_{i-1}} \right\|_2^2$
		\State $ \tilde {\vec{v}} _i = \hat {\vec{x}} _{i} + \vec{u} _i $
		\State $\hat {\vec{v}}_i = \text{Denoise}_{\beta/\lambda} \left( \tilde {\vec{v}} _i \right)$
		\State $ \vec{u}_{i+1} = \vec{u}_i + \left( {\hat{\vec{x}}}_i - {\hat{\vec{v}}_i } \right)$
		\State $ i \gets i + 1$
		\Until{stopping criterion is satisfied}
\end{algorithmic}
\end{algorithm}

\section{Linear Approximation -- A Closer Look}
The wide variety of lossy-compression methods yield a range of diverse compression performances and features. 
These often rely on the fundamental procedure of quantization, which enables to trade-off representation-precision and cost (in bits). 
The quantization concept is employed in various forms, e.g., as a scalar/vector operation, using uniform/non-uniform representation levels, and also in the extreme case of thresholding where some data elements are completely discarded (and the remaining are regularly quantized).
Furthermore, the statistical properties of the data affect the quantizer performance, and indeed, the prevalent transform-coding concept first considers the data in different orthogonal basis that enables more efficient quantization.

In this section we study the linear approximation of the scalar-quantization procedure, starting at its use for a single variable and proceeding to transform coding of vectors, where the transform-domain coefficients are independently quantized.
The analysis provided here sheds some clarifying light on our linearization strategy that is generically applied in the proposed technique to more complicated compression methods.

\subsection{Local Linear-Approximation of a Quantizer}
Let us consider a general scalar quantization function $q(x)$ that maps the real-valued input $x$ onto a discrete set of real-valued representation levels.
As $q(x)$ is a non-differentiable function we examine its linear approximation around the point $x_0$ in a limited interval defined by $\delta$ as 
\begin{IEEEeqnarray}{rCl}
	\label{eq:General quantizer - linear template}
	\eta(x_0,\delta) = \left[ x_0 - \delta, x_0 + \delta \right].
	\footnote{Here we mathematically study the problem for a given quantization function that is used for calculating the approximation, and therefore a single $\delta$ value is sufficient. However, in our generic algorithm we empirically utilize a set of $\delta$ values in Equation (\ref{eq:Jacobian estimation - k-th column}) since the compression function is unknown.}
\end{IEEEeqnarray}	
The studied approximation takes the general linear form of
\begin{IEEEeqnarray}{rCl}
	\label{eq:General quantizer - linear template}
	{ \tilde{q} }\left( x \right) = ax + b
\end{IEEEeqnarray}
where $a$ and $b$ are the linearization parameters.
The approximation in (\ref{eq:General quantizer - linear template}) introduces an error that can be measured via the Mean-Squared-Error (MSE) over the local interval $\eta(x_0,\delta)$ \footnote{We study the linearization error as a function of the approximation-interval size, which is determined by $\delta$. Clearly, by setting a sufficiently small $\delta$ we get a zero approximation-error as we shall see hereafter. However, note that the linearization error is not the only factor to consider for the selection of $\delta$.
Therefore, we should bear in mind throughout the following derivation that we do not present here an explicit method for selecting the value of $\delta$ but an analysis of the local approximation-error of the quantizer as a function of $\delta$. Nevertheless, this mathematical analysis demonstrates the important principles of linear approximation of quantizers and motivates the algorithmic design and experimental settings that are presented in the following sections.}:
\begin{IEEEeqnarray}{rCl}
	\label{eq:General quantizer - approximation error}	
&&LMSE^{SQ}\bigl( {a,b} ; \eta\left(x_0,\delta\right) \bigr)  \triangleq
\\ \nonumber
&&\qquad \frac{1}{{2\delta }}\mathop \int \limits_{{x_0} - \delta }^{{x_0} + \delta } {\bigl( {q\left( x \right) - {\tilde q}\left( x \right)} \bigr)^2}dx
\end{IEEEeqnarray}
Substituting (\ref{eq:General quantizer - linear template}) in (\ref{eq:General quantizer - approximation error}), then demanding parameter optimality by
\begin{IEEEeqnarray}{rCl}
	\label{eq:Two level quantizer - optimality demand}	
	\frac{\partial}{{\partial a}}LMSE^{SQ}\bigl( {a,b} ; \eta\left(x_0,\delta\right) \bigr) & = & 0 
	\\ \nonumber
	  \frac{\partial}{{\partial b}}LMSE^{SQ}\bigl( {a,b} ; \eta\left(x_0,\delta\right) \bigr) & = & 0,
\end{IEEEeqnarray}
leads to the following optimal parameters:
\begin{IEEEeqnarray}{rCl}
	\label{eq:Equation system of linearization paramter optimality}	
	\label{eq:Equation system of linearization paramter optimality - a}	
a^* &=& \frac{3}{{{\delta ^2}}}\left( {{L_a} - {L_b}{x_0}} \right)
\\
	\label{eq:Equation system of linearization paramter optimality - b}	
b^* &=& {L_b} - \frac{{3{x_0}}}{{{\delta ^2}}}\left( {{L_a} - {L_b}{x_0}} \right)
\end{IEEEeqnarray}
where we defined 
\begin{IEEEeqnarray}{rCl}
	\label{eq:Equation system - optimal integral defintion - a}	
	L_a &\triangleq& \frac{1}{{2\delta }}\mathop \int \limits_{{x_0} - \delta }^{{x_0} + \delta } xq\left( x \right)dx
	\\
	\label{eq:Equation system - optimal integral defintion - b}	
	L_b &\triangleq& \frac{1}{{2\delta }}\mathop \int \limits_{{x_0} - \delta }^{{x_0} + \delta } q\left( x \right)dx .
\end{IEEEeqnarray}
To better understand the values of these parameters, we shall consider several simple cases.

\subsection{The Case of Two-Level Quantization}
\label{subsec:The Case of Two-Level Quantization}
\begin{figure}
	\centering
	{\subfloat[Two-level]{\includegraphics[width=0.23\textwidth]{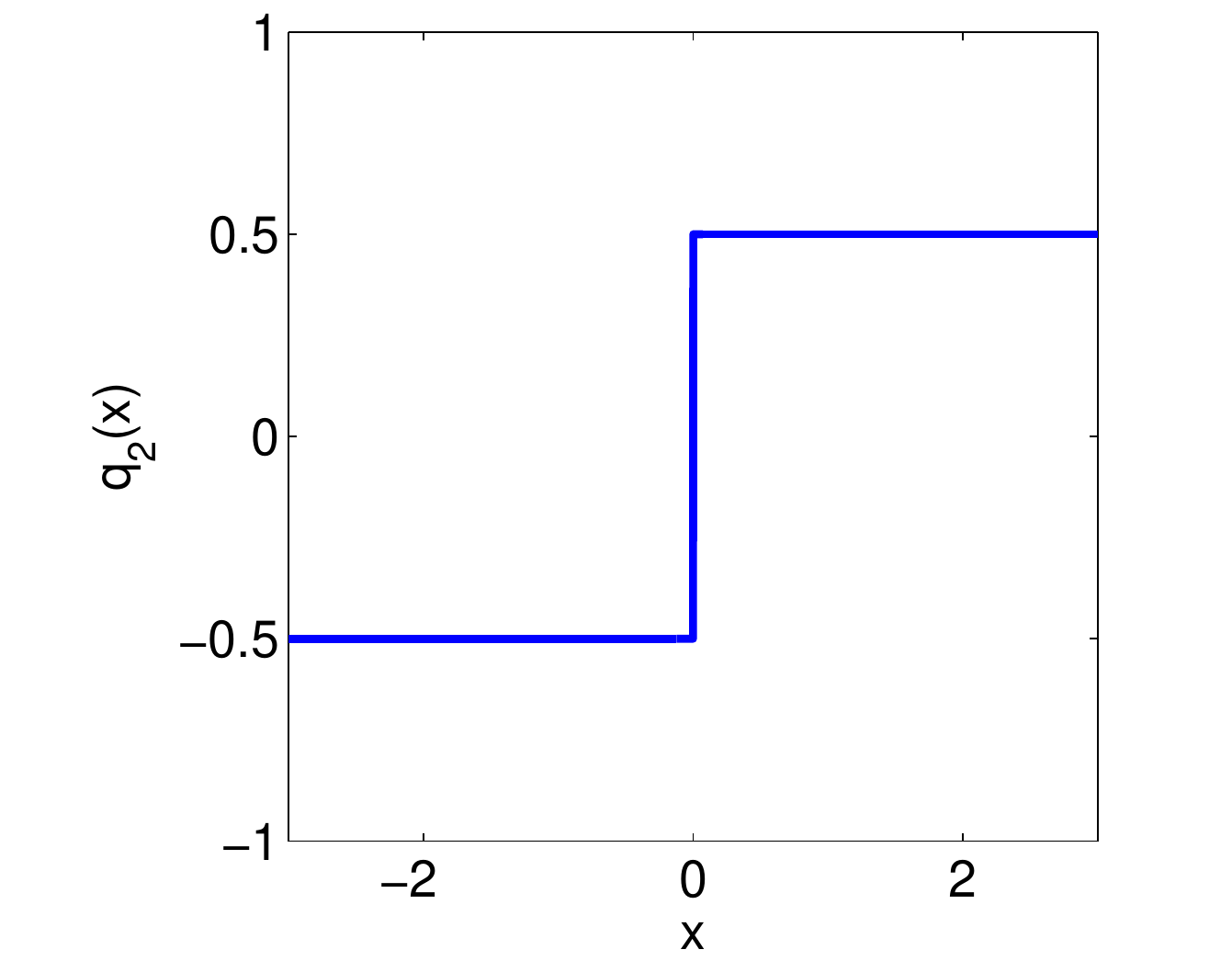} %
	\label{Fig:Two-level quantizer}} }
	{\subfloat[Uniform]{\includegraphics[width=0.22\textwidth]{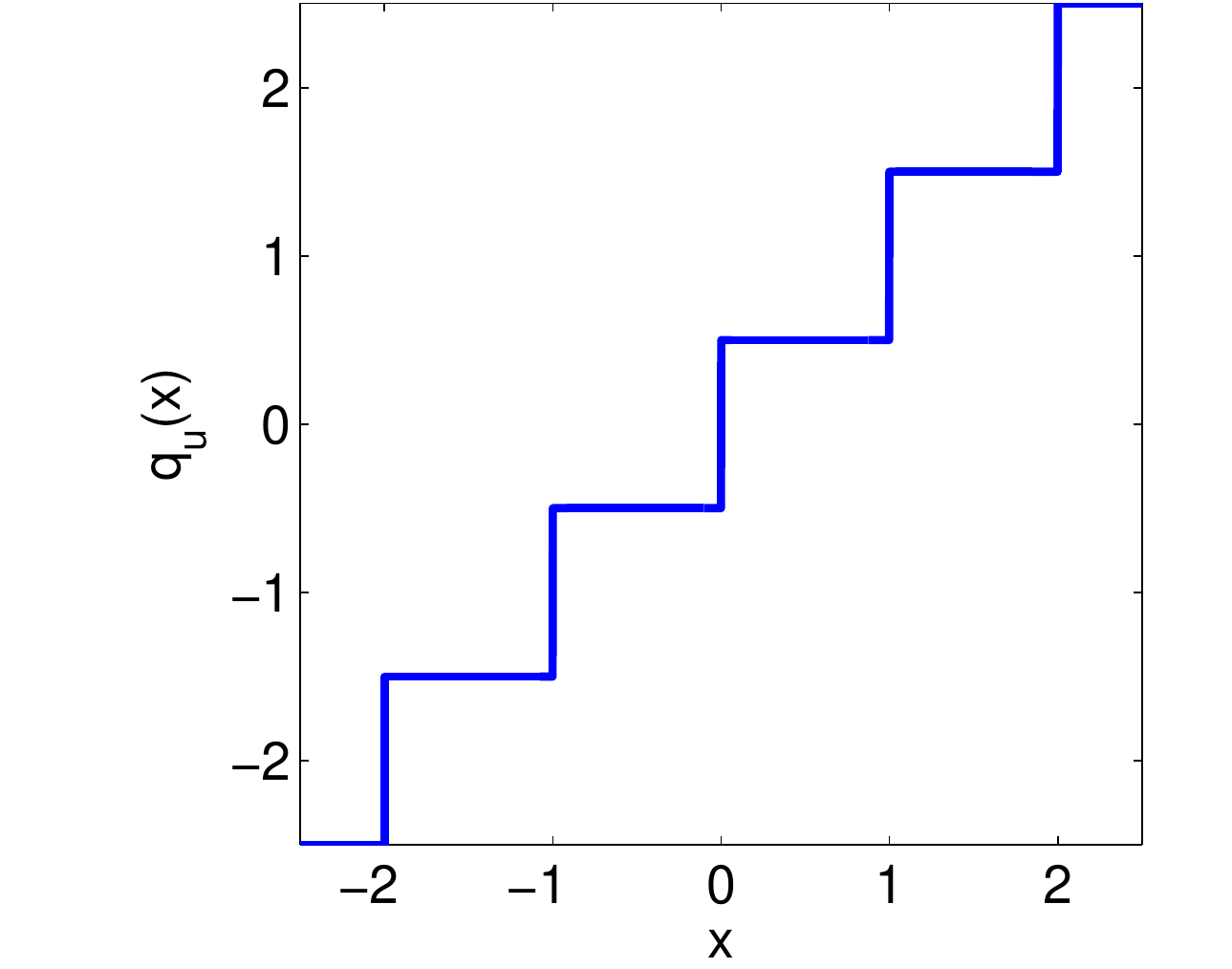} %
			\label{Fig:Uniform quantizer}}}
	\caption{Examples of scalar quantizers.} 
\end{figure}

We start by studying the elementary two-level quantizer that takes the form of a step function (Fig. \ref{Fig:Two-level quantizer}) as follows:
\begin{IEEEeqnarray}{rCl}
	\label{eq:Two level quantizer}
	{q_2 }\left( x \right) = \left\{ {\begin{array}{*{20}{c}}
			-1/2&{,{\text{for}}\,\,x \leqslant 0 } \\ 
			~~1/2 &{,{\text{for}}\,\,x > 0 } 
		\end{array}} \right.
	\end{IEEEeqnarray}
where the two output levels, $r_0=\nolinebreak -1/2$ and $r_1=\nolinebreak 1/2$, are assigned according to the input sign.
This canonic form is useful to our discussion here, since it is an asymmetric function around the origin, and thus, will simplify the mathematical analysis. Nevertheless, the form in (\ref{eq:Two level quantizer}) can be extended to any two-level quantizer using shifts and scaling that adjust the step-location and the two representation levels. Accordingly, the results in this section are easily extended, e.g., by considering the quadratic effect of the step-size scaling on the local MSE.

When the local interval is completely contained within a single decision region, i.e. $\eta(x_0,\delta) \subset \nolinebreak \left[ -\infty, 0 \right]$ or $\eta(x_0,\delta) \subset \nolinebreak \left( 0, \infty \right]$, then ${q_2 }\left( x \right)$ is locally fixed on $r_0$ or $r_1$, respectively, and therefore 
\begin{IEEEeqnarray}{rCl}
	\label{eq:Two level quantizer - linearization paramter optimality - integrals for case of one representation level - a}	
	L_a &=& \frac{1}{{2\delta }}\mathop \int \limits_{{x_0} - \delta }^{{x_0} + \delta } x r_i dx =  {x_0}r_i
	\\
	\label{eq:Two level quantizer - linearization paramter optimality - integrals for case of one representation level - b}
	L_b &=& \frac{1}{{2\delta }}\mathop \int \limits_{{x_0} - \delta }^{{x_0} + \delta } r_i dx = r_i
\end{IEEEeqnarray}
for the respective $i\in \left\lbrace 0,1 \right\rbrace$.
Then setting (\ref{eq:Two level quantizer - linearization paramter optimality - integrals for case of one representation level - a}) and (\ref{eq:Two level quantizer - linearization paramter optimality - integrals for case of one representation level - b}) in (\ref{eq:Equation system of linearization paramter optimality - a}) and (\ref{eq:Equation system of linearization paramter optimality - b}), respectively, induces the optimal values $a^*=0$ and $b^*=r_i$, that of course accurately represent the locally flat function with a corresponding zero local-MSE.

\begin{figure*}[]
 	\centering
 	 	\subfloat[$a^*$]{\includegraphics[width=0.33\textwidth]{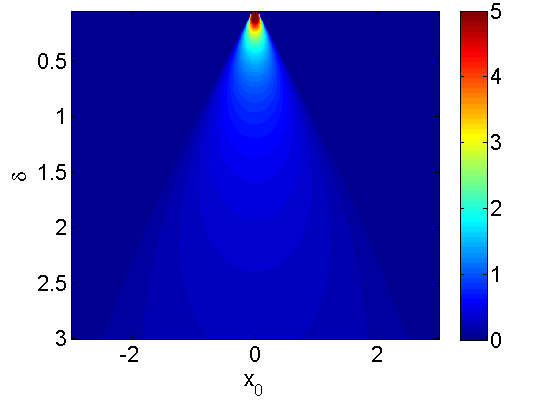}
 	 		\label{Fig:Two-level quantizer - optimal linearization - a}
 	 	}
 	 	\subfloat[$b^*$]{\includegraphics[width=0.33\textwidth]{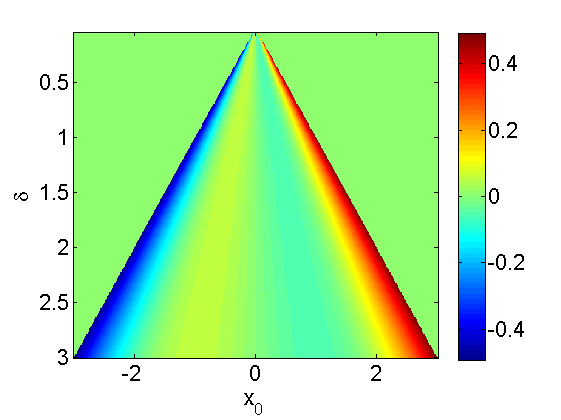}%
		\label{Fig:Two-level quantizer - optimal linearization - b}
 	 	}
 	 	\subfloat[Local MSE]{\includegraphics[width=0.33\textwidth]{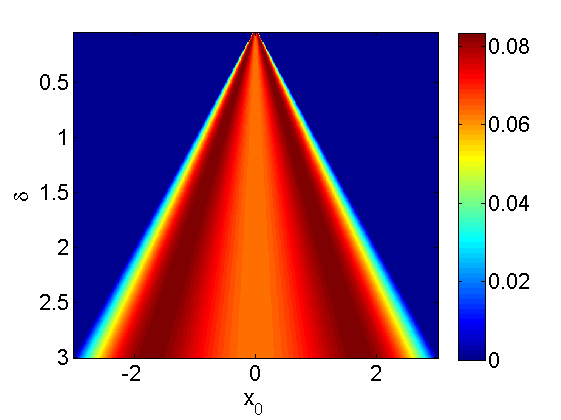}%
		\label{Fig:Two-level quantizer - optimal linearization - MSE}
 	 	}
 	\caption{Optimal linearization of a normalized two-level quantizer as function of the local interval center ($x_0$) and length ($\delta$). } 
 	\label{Fig:Two-level quantizer - optimal linearization}
 \end{figure*}

Now we turn to the more interesting case where the local interval spans over the two decision regions, i.e., $x_0-\nolinebreak\delta<\nolinebreak0$ and $x_0+\nolinebreak\delta>\nolinebreak0$. Calculating again the optimal parameter set (\ref{eq:Equation system of linearization paramter optimality - a})-(\ref{eq:Equation system of linearization paramter optimality - b}) for this scenario requires to decompose the integrals (\ref{eq:Equation system - optimal integral defintion - a})-(\ref{eq:Equation system - optimal integral defintion - b}) to the two decision regions, yielding the following optimal linearization parameters:
\begin{IEEEeqnarray}{rCl}
	\label{eq:Two level quantizer - linearization paramter optimality - case of two representation level}	
	\label{eq:Two level quantizer - linearization paramter optimality - case of two representation level - a}	
a^*&=&\frac{3}{4\delta} \left(1-\left(\frac{x_0}{\delta}\right) ^2\right)
\\
	\label{eq:Two level quantizer - linearization paramter optimality - case of two representation level - b}	
b^*&=& \frac{3 {x_0}}{4\delta} \left(  \left( \frac{x_0}{\delta} \right) ^2   - \frac{1}{3}  \right)
\end{IEEEeqnarray}
and, using (\ref{eq:General quantizer - approximation error}), the corresponding error (for $\delta>\lvert x_0 \rvert$) is\footnote{Recall that for $\delta<\lvert x_0 \rvert$ the approximation local-MSE is zero.}
\begin{IEEEeqnarray}{rCl}
	\label{eq:Two level quantizer - optimal linearization error}	
	&& LMSE^{SQ}\bigl( {a^*,b^*} ; \eta\left(x_0,\delta\right) \bigr) = 
	\\ \nonumber 
	&&\qquad \frac{1}{16} \left( 1 + 3 \left(\frac{x_0}{\delta}\right)^2 \right) \left( 1 - \left(\frac{x_0}{\delta}\right)^2 \right).
\end{IEEEeqnarray}
One should note that on the limit of the global linear approximation, i.e., when $\delta \rightarrow \infty$, the optimal parameters are
\begin{IEEEeqnarray}{rCl}
\label{eq:Two level quantizer - linearization paramter optimality - case of two representation level - global case - a}
	\lim\limits_{\delta \rightarrow \infty}a^* &=& 0
	\\
		\label{eq:Two level quantizer - linearization paramter optimality - case of two representation level - global case - b}
	\lim\limits_{\delta \rightarrow \infty}b^* &=& 0.
\end{IEEEeqnarray}
This asymptotic fitting to a constant-valued function is also expressed in the numerical results in Fig. \ref{Fig:Two-level quantizer - optimal linearization - a}-\ref{Fig:Two-level quantizer - optimal linearization - b}. 

Let us study the optimal approximation for the non-trivial case of $\delta>\lvert x_0 \rvert$. 
First, we notice that the error tends to zero as $\delta$ gets closer to $\lvert x_0 \rvert$. Second, The maximal error is obtained for $\delta = \sqrt{3} \lvert x_0 \rvert$ and its value is $\frac{1}{12}$.
Moreover, for approximation around the non-differentiable point, i.e. $x_0 = 0$, the error is a constant and, therefore, independent of $\delta$. 
This interesting observation is a special case of a more general behavior where a constant error value is achieved for any $\left(x_0, \delta\right)$ pair that is on the line $\delta = c \lvert x_0 \rvert$ for some $c \in \left[0,\infty\right)$. This constant local-MSE is due to the fixed ratio between the lengths of the subintervals $\left[x_0 - \delta, 0\right]$ and $\left[0, x_0 + \delta\right]$, determining the optimal approximation in this case.
The latter analysis is clearly exhibited in the numerical results in Fig. \ref{Fig:Two-level quantizer - optimal linearization}.

The numerical results also demonstrate the following behavior of the approximation as function of $\delta$. At the beginning, the solution gradually considers the step by having an increasingly steeper slope, then, the approximation begins to approach the asymptotic solution of a flat line.
It is also observed that the approximation is useful (in terms of relatively low error) when the interval size tends to be the minimal that contains the discontinuity point, located here at $0$. 
Furthermore, in some sense, finding the best interval for approximating around $x_0 \ne 0$ is like measuring the distance of $x_0$ from the step.

\subsection{The Case of Multi-Level Uniform Quantization}
\label{subsec:The Case of Multi-Level Uniform Quantization}
Let us extend the above analysis to a multi-level uniform quantizer in the mid-riser form \cite[p. 137]{gersho2012vector} (Fig. \ref{Fig:Uniform quantizer}):
\begin{IEEEeqnarray}{rCl}
	\label{eq:uniform quantizer formula}
	q_u\left( {{x}} \right) =  {\left\lfloor {x} \right\rfloor  + \frac{1}{2}} .
\end{IEEEeqnarray}
Here the quantization step is of unit length, and accordingly the $i^{th}$ decision region, $\left[ d_i,d_{i+1} \right) = \left[ i, i+1\right)$, maps the input to the $i^{th}$ representation level $r_i = i + \frac{1}{2}$. Note that $i$ is an integer that may be positive or negative.
As in the previous case, this normalized quantizer form yields a simplified analysis that is, however, extendable to any uniform quantizer by shifts and scaling.

The optimal local linear approximation for this uniform quantizer is obtained by calculating  (\ref{eq:Equation system of linearization paramter optimality - a})-(\ref{eq:Equation system of linearization paramter optimality - b}) for the formula in (\ref{eq:uniform quantizer formula}). 
Again, the solution depends on the interval layout. In the simplest case, the considered interval is completely contained within a single decision region, i.e., $\eta(x_0,\delta) \subset \nolinebreak \left[ d_i, d_{i+1} \right]$ for some $i$. Here $q\left(x\right)=r_i$ for any $x \in \eta(x_0,\delta)$. 
Clearly, the corresponding discussion for the two-level quantizer (see section \ref{subsec:The Case of Two-Level Quantization}) also holds here, meaning that $a^*=0$ and $b^*=r_i$ with a zero local-MSE.

Another scenario that coincides with the two-level quantizer is when $ x_0 - \delta \in \left[ d_{i-1}, d_{i} \right]$ and $ x_0 + \delta \in \left[ d_i, d_{i+1} \right]$, i.e., the interval is spread over only two adjacent decision regions. Indeed, the optimal parameters here are obtained by appropriately shifting the results in (\ref{eq:Two level quantizer - linearization paramter optimality - case of two representation level - a})-(\ref{eq:Two level quantizer - linearization paramter optimality - case of two representation level - b}). However, note that the multi-level quantizer has two levels only for $\delta <\nolinebreak min\left\lbrace d_{i+1} - x_0 , x_0 - d_{i-1} \right\rbrace <\nolinebreak 1$.

\begin{figure*}[]
 	\centering
 	\subfloat[$a_u^*$]{\label{Fig:Uniform quantizer - optimal linearization - a}
 		\includegraphics[width=0.33\textwidth]{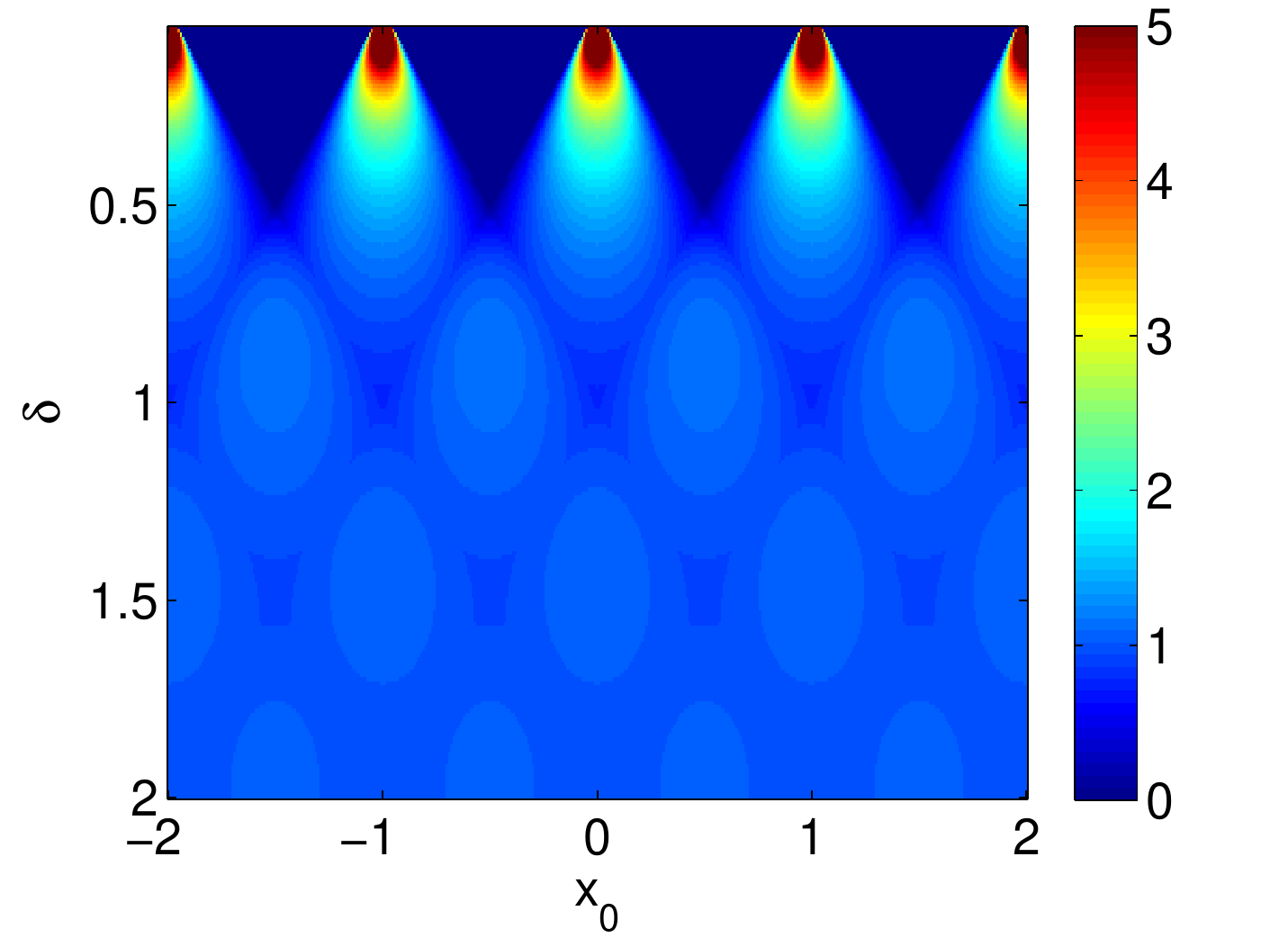}
 	}
 	\subfloat[$b_u^*$]{\label{Fig:Uniform quantizer - optimal linearization - b}
 		\includegraphics[width=0.33\textwidth]{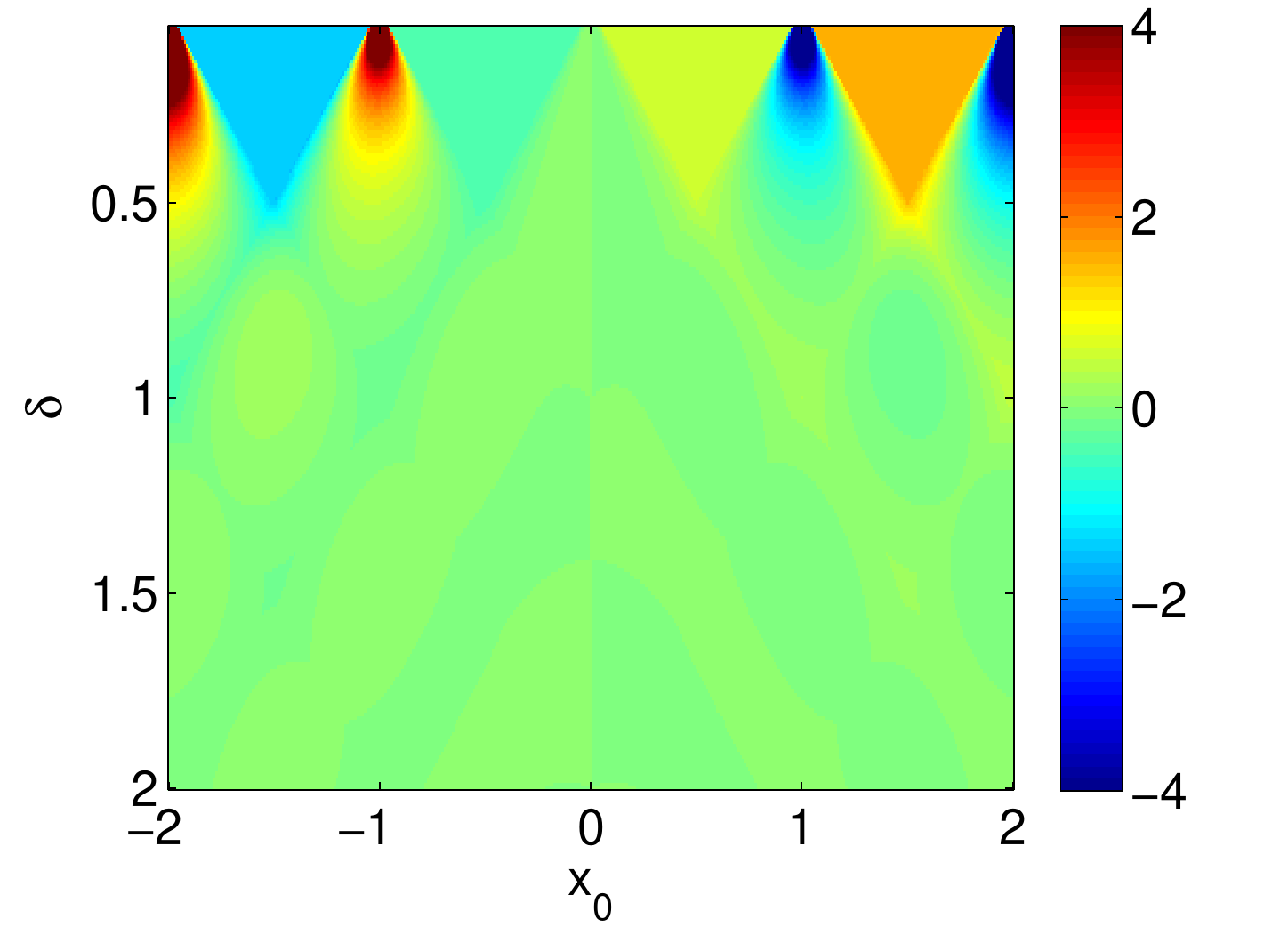}%
 	}
 	\subfloat[Local MSE] {\label{Fig:Uniform quantizer - optimal linearization - MSE}
 		\includegraphics[width=0.33\textwidth]{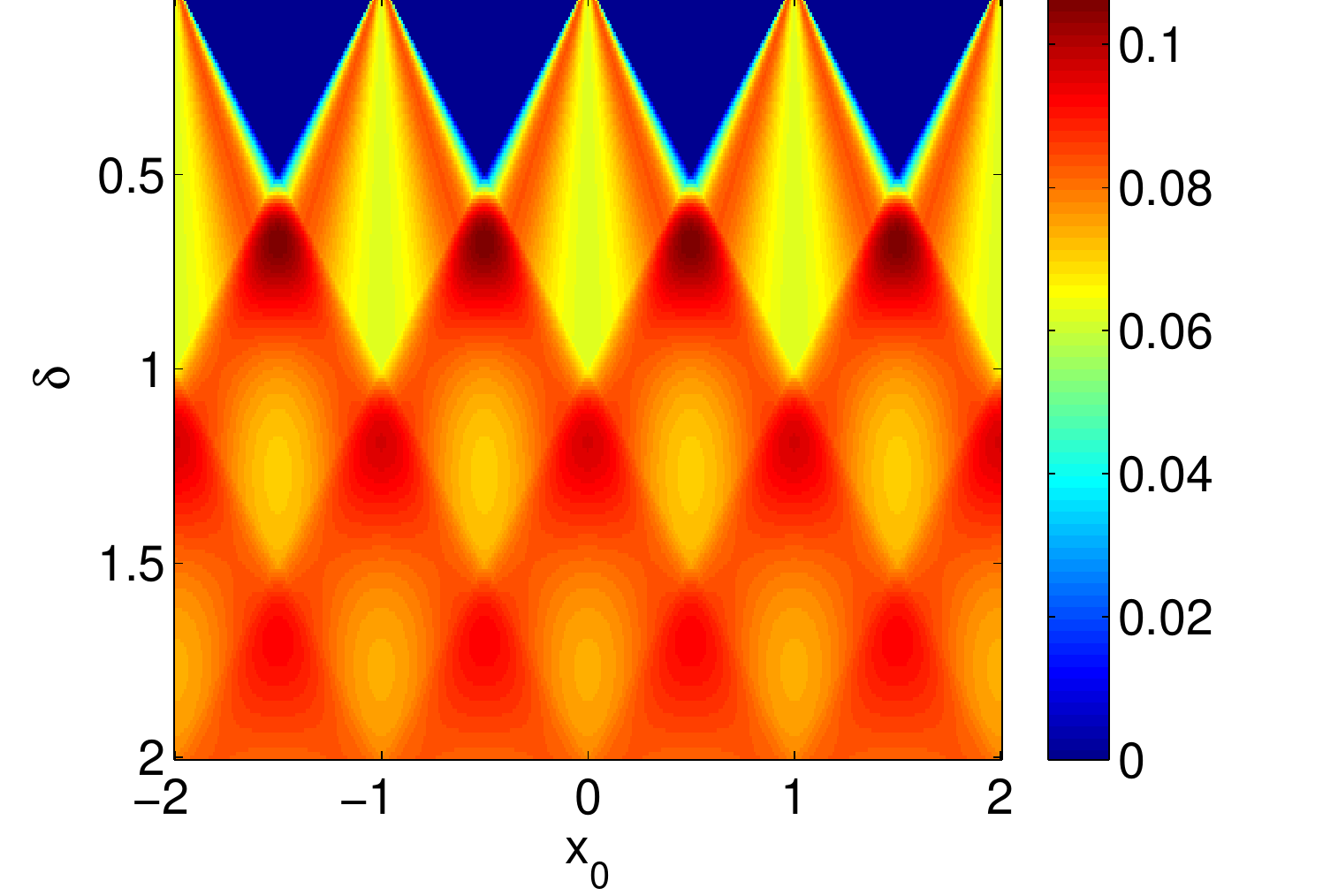}%
 	}
 	\caption{Optimal linearization of a normalized uniform quantizer as function of the local interval center ($x_0$) and length ($\delta$). } 
 	\label{Fig:Uniform quantizer - optimal linearization}
 \end{figure*}

Now we proceed to the main case, where the approximation interval spans over more than two decision regions, i.e., $ x_0 - \nobreak \delta \in \left[ d_{i}, d_{i+1} \right]$ and $ x_0 + \delta \in \left[ d_j, d_{j+1} \right]$ for $j-i>1$. 
First, we express the uniform quantization function as a sum of shifted two-level quantizers:
\begin{IEEEeqnarray}{rCl}
	\label{eq:uniform quantizer as a sum of two-level quantizers}
	q_u\left( {{x}} \right) =  \sum\limits_{\tau=-\infty}^{\infty} q_2(x-\tau),
\end{IEEEeqnarray}
where $q_2(\cdot)$ was defined in (\ref{eq:Two level quantizer}).
Then, using (\ref{eq:uniform quantizer as a sum of two-level quantizers}) we can develop (\ref{eq:Equation system - optimal integral defintion - a})-(\ref{eq:Equation system - optimal integral defintion - b}) to the following forms:
\begin{IEEEeqnarray}{rCl}
	\label{eq:uniform quantizer - La decomposed}
	L_a^u =  \sum\limits_{\tau=-\infty}^{\infty} L_a^{\tau}
	\\
	\label{eq:uniform quantizer - Lb decomposed}
	L_b^u =  \sum\limits_{\tau=-\infty}^{\infty} L_b^{\tau},
\end{IEEEeqnarray}
where $L_a^{\tau}$ and $L_b^{\tau}$ are the corresponding values for the two-level quantizer $q_2(x-\tau)$. 
These allow us to write the optimal linearization parameters of the uniform quantizer as the summation of the optimal parameters of the shifted two-level quantizers, namely
\begin{IEEEeqnarray}{rCl}
	\label{eq:uniform quantizer - a decomposed}
	&& a^*_u =  \sum\limits_{\tau=-\infty}^{\infty} a^*_{\tau}
	\\
	\label{eq:uniform quantizer - b decomposed}
	&& b^*_u =  \sum\limits_{\tau=-\infty}^{\infty} b^*_{\tau},						
\end{IEEEeqnarray}
where $a^*_{\tau}$ and $b^*_{\tau}$ are the optimal linearization parameters for $q_2(x-\tau)$ and are obtainable by shifting the expressions in (\ref{eq:Two level quantizer - linearization paramter optimality - case of two representation level - a})-(\ref{eq:Two level quantizer - linearization paramter optimality - case of two representation level - b}).
This analytic relation between the linearization of the uniform and the two-level quantizers is clearly exhibited in the numerical results (see Fig. \ref{Fig:Uniform quantizer - optimal linearization}) in the form of a periodic structure.

The numerical calculations (Fig.  \ref{Fig:Uniform quantizer - optimal linearization}) also show convergence to the global approximation parameters
\begin{IEEEeqnarray}{rCl}
	\label{eq:Uniform quantizer - linearization paramter optimality - case of two representation level - global case - a}
	\lim\limits_{\delta \rightarrow \infty}a^*_u &=& 1
	\\
	\label{eq:Uniform quantizer - linearization paramter optimality - case of two representation level - global case - b}
	\lim\limits_{\delta \rightarrow \infty}b^*_u &=& 0,
\end{IEEEeqnarray}
which imply $\lim\limits_{\delta \rightarrow \infty}  C_{lin} \left(x\right) = x$.
In order to explain the results in Fig. \ref{Fig:Uniform quantizer - optimal linearization}, we return to the interpretation of a multi-level quantizer as a sum of shifted two-level quantizers (as expressed in Eq. (\ref{eq:uniform quantizer as a sum of two-level quantizers})). 
First, examining the case of approximation around a decision level, shows that at each point of $\delta = k$ (for integer $k$ values), two additional representation levels are included in the approximation (one on each side of the interval) and affect the optimal approximation.
Comparing Fig. \ref{Fig:Uniform quantizer - optimal linearization} to Fig. \ref{Fig:Two-level quantizer - optimal linearization} reveals that the effect of each of these added representation levels is like approximating a two-level quantizer around a point that differs from its threshold level.
Evaluating the approximation around a point that is not a decision level (see Fig. \ref{Fig:Uniform quantizer - optimal linearization} while considering non-integer $x_0$ values) extends the previous behavior by combining two unsynchronized periodic patterns, each of them stems from a recurrent addition of representation levels from a different side.

The MSE plot (Fig. \ref{Fig:Uniform quantizer - optimal linearization - MSE}) 
shows that, for nontrivial intervals that contain at least one non-differentiable point, the minimal MSE is obtained for approximation over a small interval that includes only the nearest decision level.
This somewhat resembles the underlying principle of the dithering procedure \cite{schuchman1964dither}, where the points within a quantization-cell are differentiated by an added noise that statistically maps them to neighboring cells according to their relative proximity.
Moreover, maximal MSE of 0.106 is obtained for $\delta=0.67$ and $x_0 = \frac{1}{2} + i$ (for $i=0, \pm 1, \pm 2, ...$), where only the two adjacent non-differentiable points affect the linearization. This can also be shown analytically by setting the decompositions in (\ref{eq:uniform quantizer as a sum of two-level quantizers}) and (\ref{eq:uniform quantizer - a decomposed})-(\ref{eq:uniform quantizer - b decomposed}) into (\ref{eq:General quantizer - approximation error}), resulting in 
\begin{IEEEeqnarray}{rCl}
	\label{eq:Uniform quantizer - decomposition of LMSE}
	&& LMSE^{SQ}_u\bigl( {a^*,b^*} ; \eta\left(x_0,\delta\right) \bigr) =
	\\ \nonumber
	&&\qquad 
	\sum\limits_{\tau=-\infty}^{\infty} LMSE^{SQ}_{\tau}\bigl( {a^*_{\tau},b^*_{\tau}} ; \eta\left(x_0,\delta\right) \bigr)
	\\ \nonumber
	&&\qquad + \frac{1}{{2\delta }}\sum\limits_{\substack{\tau, \nu=-\infty\\\tau\ne\nu}}^{\infty} \mathop \int \limits_{{x_0} - \delta }^{{x_0} + \delta } {\bigl( {q_2(x-\tau) - a^*_{\tau}x - b^*_{\tau}} \bigr)} \times
	\\ \nonumber
	&&\qquad\qquad\qquad\qquad\qquad\times{\bigl( {q_2(x-\nu) - a^*_{\nu}x - b^*_{\nu} } \bigr)} dx,
\end{IEEEeqnarray}
where $LMSE^{SQ}_{\tau}\bigl( {a^*_{\tau},b^*_{\tau}} ; \eta\left(x_0,\delta\right) \bigr)$ is the optimal LMSE for $q_2(x-\tau)$ as available by shifting the expression in (\ref{eq:Two level quantizer - optimal linearization error}).


\subsection{Transform Coding}
\label{subsec:Linearization analysis for transform coding}
We now turn to generalize the discussion to compression of multidimensional signals by considering the widely used concept of transform coding, where scalar quantization is applied in the transform domain.
We examine coding of an $N$-length signal vector using a unitary transform, that can be formulated as the vector-valued function
\begin{IEEEeqnarray}{rCl}
	\label{eq:General tranform coding - template}
	 C\left( \vec{x} \right) = \mtx{U}Q\left(  \mtx{U}^T \vec{x}  \right) ,
\end{IEEEeqnarray}
where $\vec{x}$ is the $N\times 1$ signal to compress, $\mtx{U}$ is an $N\times N$ unitary matrix, and $Q\left(\cdot\right)$ is a vector-valued quantization function that scalarly quantizes the input components, i.e.,
\begin{IEEEeqnarray}{rCl}
	\label{eq:scalar quantization of a vector }
	Q\left(  \vec x  \right) = \left[ {\begin{array}{*{20}{c}}
		{q\left( {{x_1}} \right)} \\ 
		\vdots  \\ 
		{q\left( {{x_N}} \right)} 
		\end{array}} \right]
\end{IEEEeqnarray}
where $q\left(\cdot\right)$ is a single-variable scalar quantization function as studied above, and $x_i$ is the $i^{th}$ component of the vector $\vec{x}$. Moreover, as the last definition exhibits, the discussion is simplified by assuming identical quantization rules to all vector components.


As scalar quantization is a building block of the transform coding procedure (\ref{eq:General tranform coding - template}), it imposes its non-differentiable nature on $C(\vec{x})$.
Let us consider the linear approximation of $C(\vec{x})$ around the point $\vec x_0 \in \mathbb{R} ^N$ in a limited neighborhood of a high-dimensional cube defined by $\delta$ as 
\begin{IEEEeqnarray}{rCl}
	\label{eq:General tranform coding - local approximation area}
	&& \eta(\vec{x_0},\delta) = \left\lbrace \vec{x} ~\vert \left\| \vec{x} - \vec{x_0} \|_{\infty} \le \delta \right. \right\rbrace.
\end{IEEEeqnarray}	
The approximation takes the general multidimensional linear form of
\begin{IEEEeqnarray}{rCl}
	\label{eq:General transform coding - linear template}
	{ \tilde{C} }\left( \vec x \right) = \mtx{A} \vec{x} + \vec{b}
\end{IEEEeqnarray}
where $\mtx A \in \mathbb{R}^{N \times N}$ and $\vec b \in\mathbb{R}^{N}$ are the linearization parameters.
The local MSE of approximating the transform-coding procedure around $\vec x_0$ is defined as 
\begin{IEEEeqnarray}{rCl}
	\label{eq:General transform coding - approximation error}
	&& LMSE^{TC}\bigl( {\mtx{A},\vec{b}} ;{\eta(\vec{x_0},\delta)} \bigr) \triangleq
	\\ \nonumber 
	&& \qquad \frac{1}{{ \lvert \eta(\vec{x_0},\delta) \rvert }}\mathop \int \limits_{\eta(\vec{x_0},\delta)  } { \left\| {C(\vec{x}) - { \tilde{C} }\left( \vec x \right) } \right\| ^2 _2}d\vec{x}
\end{IEEEeqnarray}
By substituting (\ref{eq:General transform coding - linear template}) in (\ref{eq:General transform coding - approximation error}) and using the energy-preservation property of unitary transforms, we get the equivalent error expression in the transform-domain 
\begin{IEEEeqnarray}{rCl}
	\label{eq:General transform coding - approximation error - transform domain}
	&& \frac{1}{{ \lvert \transU{\eta}(\vec{x}_0,\delta) \rvert }}\mathop \int \limits_{\transU{\eta} ({\vec{x}}_0,\delta)  } { \left\| Q(\transU{\vec{x}}) - \transU{\mtx{A}} \transU{\vec{x}} - \transU{\vec{b}} \right\| ^2 _2}d \transU{\vec{x}}
\end{IEEEeqnarray}
where
\begin{IEEEeqnarray}{rCl}
	\label{eq:General transform coding - approximation error - transform domain - definitions - x}
	\transU{\vec{x}} & = & \mtx{U}^T \vec{x}
	\\ \nonumber
	\label{eq:General transform coding - approximation error - transform domain - definitions - x0}
	\transU{\vec{x}}_0 & = & \mtx{U}^T \vec{x}_0
	\\ \nonumber
	\label{eq:General transform coding - approximation error - transform domain - definitions - A}
	\transU{\mtx{A}} & = & \mtx{U}^T \mtx{A} \mtx{U}
	\\ \nonumber
	\label{eq:General transform coding - approximation error - transform domain - definitions - b}
	\transU{\vec{b}} & = & \mtx{U}^T \vec{b}
	\\ \nonumber
	\label{eq:General transform coding - approximation error - transform domain - definitions - area size preservation}
	\lvert \transU{\eta}(\vec{x}_0,\delta) \rvert & = & \lvert {\eta}(\vec{x}_0,\delta) \rvert .
\end{IEEEeqnarray}
and the rotated approximation area (or volume), $\transU{\eta}$, is defined around $\transU{\vec{x}}_0$ and may not have sides that are aligned with the axes. 

Let us generally define the local linearization error for compression of a signal vector, $\vec{x}$, by identical scalar quantization of its components, $x_i$:
\begin{IEEEeqnarray}{rCl}
	\label{eq:scalar quantization of a vector - approximation error - general definition}
	&& LMSE^{SQV}\bigl( {\mtx{A},\vec{b}} ;{\bar\eta} \bigr) \triangleq
	\\ \nonumber
	&&\qquad = \frac{1}{{ \lvert \bar\eta \rvert }}\mathop \int \limits_{\bar\eta  } { \left\| Q({\vec{x}}) - {\mtx{A}} {\vec{x}} - {\vec{b}} \right\| ^2 _2}d {\vec{x}}
	\\ \nonumber
	&&\qquad = \frac{1}{{ \lvert \bar\eta \rvert }}\mathop \sum\limits_{i=1}^{N} \int \limits_{\bar\eta  } { \left( q({{x}}_i) - {\vec{a}}^T_i {\vec{x}} - {{b}}_i \right) ^2}d {\vec{x}}
\end{IEEEeqnarray}
where $\bar\eta$ is an arbitrary shaped approximation area, and the last equality relies on the separability of $Q\left(\cdot\right)$ and the $L_2$-norm definition.

Equations (\ref{eq:General transform coding - approximation error - transform domain}) and (\ref{eq:scalar quantization of a vector - approximation error - general definition}) clearly show that the MSE of approximating transform-coding (\ref{eq:General transform coding - approximation error - transform domain}) reduces to the linearization-error of scalar quantization of the transform coefficients, namely, 
\begin{IEEEeqnarray}{rCl}
	\label{eq:General transform coding - approximation error - transform domain - sum of components}
	&& LMSE^{TC}\bigl( {\mtx{A},\vec{b}} ; {\eta}(\vec{x}_0,\delta) \bigr) 
	\\ \nonumber
	&&~~~~ = LMSE^{SQV}\left( {\transU{\mtx{A}},\transU{\vec{b}}} ; \transU{\eta}(\vec{x}_0,\delta) \right) 
	\\ \nonumber
	&&~~~~ = \frac{1}{{ \lvert \transU{\eta}(\vec{x}_0,\delta) \rvert }}\mathop \sum\limits_{i=1}^{N} \int \limits_{\transU{\eta} ({\vec{x}}_0,\delta)  } { \left( q(\transU{{x}}_i) - \transU{\vec{a}}^T_i \transU{\vec{x}} - \transU{{b}}_i \right) ^2}d \transU{\vec{x}}
\end{IEEEeqnarray}
where $\transU{\vec{a}}^T _i$ is the $i^{th}$ row of $\transU{\mtx{A}}$, and $\transU{{b}}_i$ is the $i^{th}$ element of the vector $\transU{\vec{b}}$.

\begin{figure}
	\includegraphics[width=0.45\textwidth]{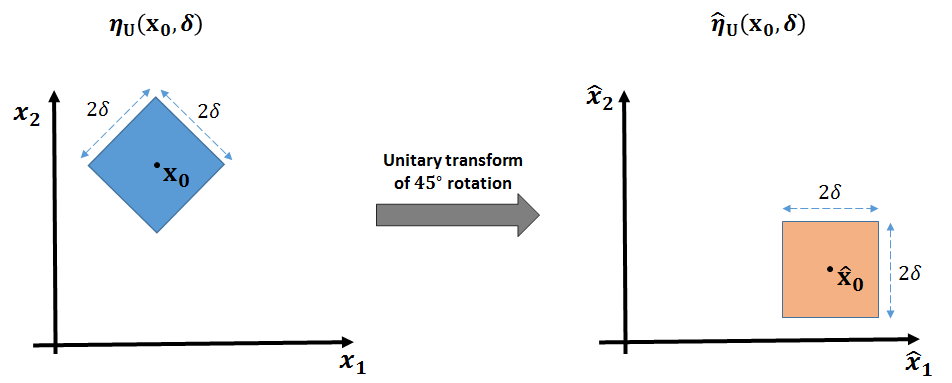} 
	\caption{Transformation of the approximation area. Exemplified in $\mathbb{R}^2$ for the unitary transform of $45^\circ$-rotation and a signal-domain area that is rotated in accordance to the transform. } 
	\label{Fig:approximation_area_transformation_rotated}		
\end{figure}

While the separability of $Q\left(\cdot\right)$ was utilized to have integrals in (\ref{eq:General transform coding - approximation error - transform domain - sum of components}) that consider quantization of single transform-coefficients, the integration is still over 
a multidimensional area that is not necessarily separable (i.e., not aligned with the axes).
We can remedy this by starting from an appropriately rotated area in the signal-domain, $\eta_U (\vec{x}_0,\delta)$, such that its transform-domain counterpart is aligned with the axes (see Fig. \ref{Fig:approximation_area_transformation_rotated}):
\begin{IEEEeqnarray}{rCl}
		\label{eq:General tranform coding - local approximation area - the transform-domain counterpart of signal-domain rotated area}
		&&\transU{\eta}_U(\vec{x}_0,\delta) = \left\lbrace \transU{\vec{x}} ~\vert \left\| \transU{\vec{x}} - \transU{\vec{x}}_0 \|_{\infty} \le \delta \right. \right\rbrace .
\end{IEEEeqnarray}
Note that  $\eta_U (\vec{x}_0,\delta)$ is not necessarily the optimally shaped approximation area as it is used here for the analytic simplicity of having full separability in the  transform domain. 
We continue our transform-domain analysis by adopting this separable integration-area.

Recall that we look for the optimal linear approximation of the signal-domain function $C\left( \vec{x} \right)$. This is obtainable by finding the optimal transform-domain parameters $\transU{\mtx{A}}^*$ and $\transU{\vec{b}}^*$ and then transforming them back to the signal domain. Following this strategy we first pose the componentwise optimality demands in the transform domain:
\begin{IEEEeqnarray}{rCl}
	\label{eq:General transform coding - optimality demand in the transform domain}
	&& \frac{\partial}{{\partial \transU{a}_{ij}}} LMSE^{SQV} \left( {\transU{\mtx{A}},\transU{\vec{b}}} ; \transU{\eta}_U (\vec{x}_0,\delta) \right)  =  0 \text{   for   } i,j=1,...,N
	\nonumber \\	
	&& \frac{\partial}{{\partial \transU{b}_i}} LMSE^{SQV} \left( {\transU{\mtx{A}},\transU{\vec{b}}} ; \transU{\eta}_U (\vec{x}_0,\delta) \right)  =  0 \text{   for   } i=1,...,N
	\nonumber \\			
\end{IEEEeqnarray}
Some calculations show that the solution satisfying the optimality conditions consists of a diagonal matrix  $\transU{\mtx{A}}^*$ (i.e., $\transU{a}^*_{ij} =0$ for $i\ne j$) such that the parameter pair $\left( \transU{a}^*_{ii} , \transU{b}^*_{i} \right)$ is the one obtained for optimal approximation of a single-variable quantizer over the interval $\left[ \transU{\vec{x}}_0^{\left( i \right)} - \delta, \transU{\vec{x}}_0^{\left( i \right)} + \delta \right]$ as generally given in (\ref{eq:Equation system of linearization paramter optimality - a})-(\ref{eq:Equation system of linearization paramter optimality - b}). Then, the signal-domain parameters are given as
\begin{IEEEeqnarray}{rCl}
	\label{eq:General transform coding - approximation error - optimality in signal-domain - A}
	\mtx{A}^* & = & \mtx{U} \transU{\mtx{A}}^* \mtx{U}^T = \sum\limits_{i=1}^{N} \transU{a}^*_{ii} \vec{u}_i \vec{u}_i^T
	\\
	\label{eq:General transform coding - approximation error - optimality in signal-domain - b}
	\vec{b}^* & = & \mtx{U} \transU{\vec{b}}^*  .
\end{IEEEeqnarray}
where the last equality in (\ref{eq:General transform coding - approximation error - optimality in signal-domain - A}) is due to the diagonality of $\transU{\mtx{A}}^*$ and $\vec{u}_i$ denotes the $i^{th}$ column of $\mtx{U}$.
The corresponding optimal error is equivalent in the signal and transform domains, hence can be expressed in a simplified form as
\begin{IEEEeqnarray}{rCl}
	\label{eq:General transform coding - approximation error - optimal}
	&& LMSE^{TC}\bigl( {\mtx{A}^*,\vec{b}^*} ; {\eta_U}(\vec{x}_0,\delta) \bigr) 
	\\ \nonumber
	&&~~~~ = LMSE^{SQV}\left( {\transU{\mtx{A}}^*,\transU{\vec{b}}^*} ; \transU{\eta}_U (\vec{x}_0,\delta) \right) 
	\\ \nonumber
	&&~~~~ = \frac{1}{{ 2 \delta }} \mathop \sum\limits_{i=1}^{N}  \int \limits_{ \transU{\vec{x}}_0^{\left( i \right)}  -  \delta  }^{\transU{\vec{x}}_0^{\left( i \right)}  + \delta } { \left( q(\transU{{x}}_i) - \transU{a}_{ii}^* \transU{{x}}_i - \transU{{b}}_i^* \right) ^2}d \transU{\vec{x}}
	\\ \nonumber
	&&~~~~ = \sum\limits_{i=1}^{N} LMSE^{SQ}\left( {\transU{a}_{ii}^*,\transU{{b}}_i^* } ; {\eta}(\transU{\vec{x}}_0^{\left( i \right)},\delta) \right) 
\end{IEEEeqnarray}
The last expression exhibits the approximation error of transform-coding as the sum of the errors of the separate linearization of the scalar quantization of the transform-domain coefficients. 
Although the assumed scenario includes equal quantization procedure for all the coefficients, the contributed errors by the various elements are different as each has its own scalar approximation-point $\transU{\vec{x}}_0^{(i)}$ located differently with respect to the quantization lattice.

\begin{figure}
	\includegraphics[width=3.1in]{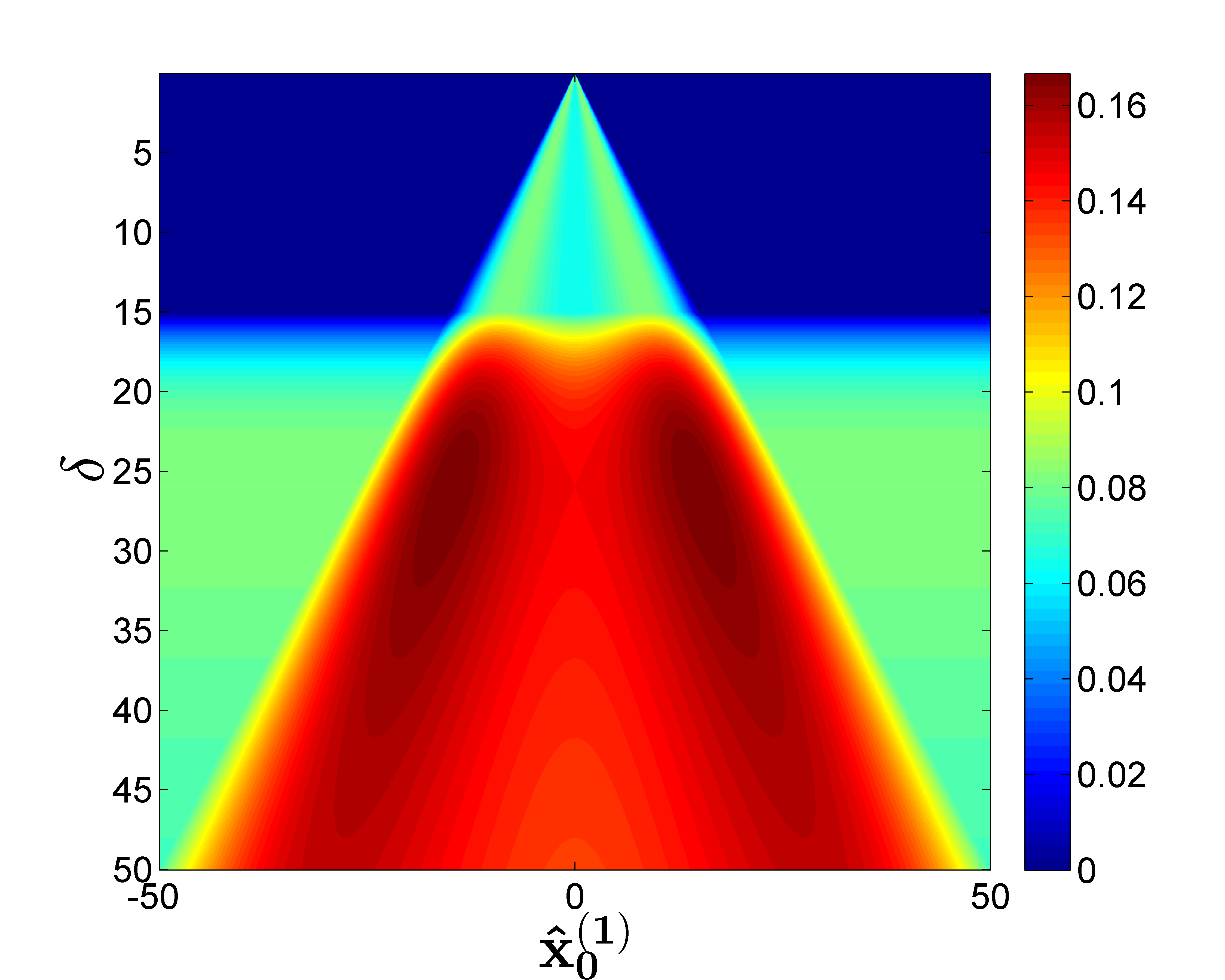}
	\caption{Overall MSE of optimal linearization of the exemplary transform coding procedure (for signals in $\mathbb{R} ^2$ and equal transform-domain quantizers).} 
	\label{Fig:Transform coding - 2x2 45degrees rotation - optimal linearization - overall MSE}
\end{figure}

\begin{figure}
	\centering
	{\subfloat[$\transU{a}_{11}^*$]{\includegraphics[width=0.23\textwidth]{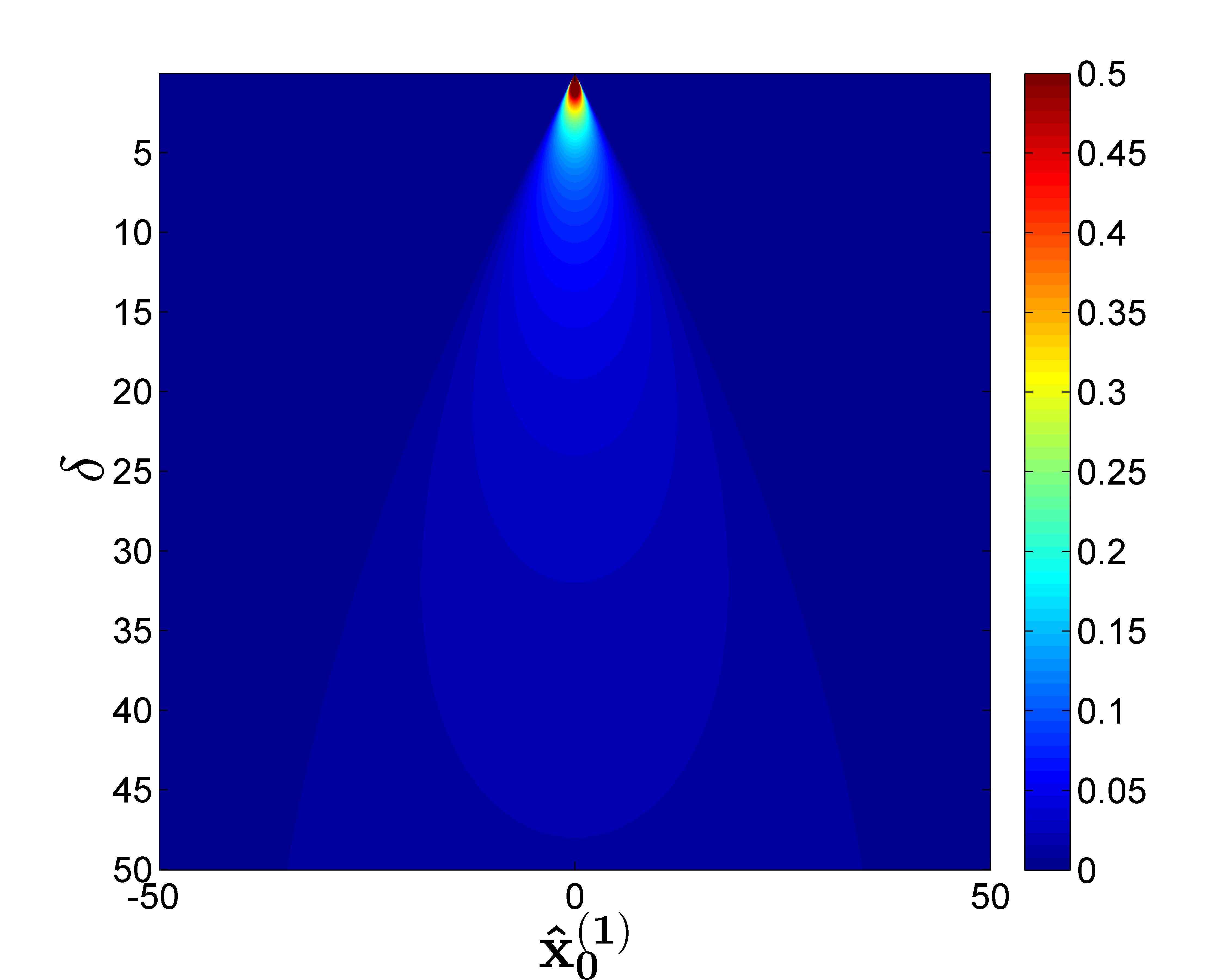} %
			\label{Fig:transform_coding__optimal_linearization__transform_domain_analysis__component_1__a11}} }
	{\subfloat[$\transU{a}_{22}^*$]{\includegraphics[width=0.23\textwidth]{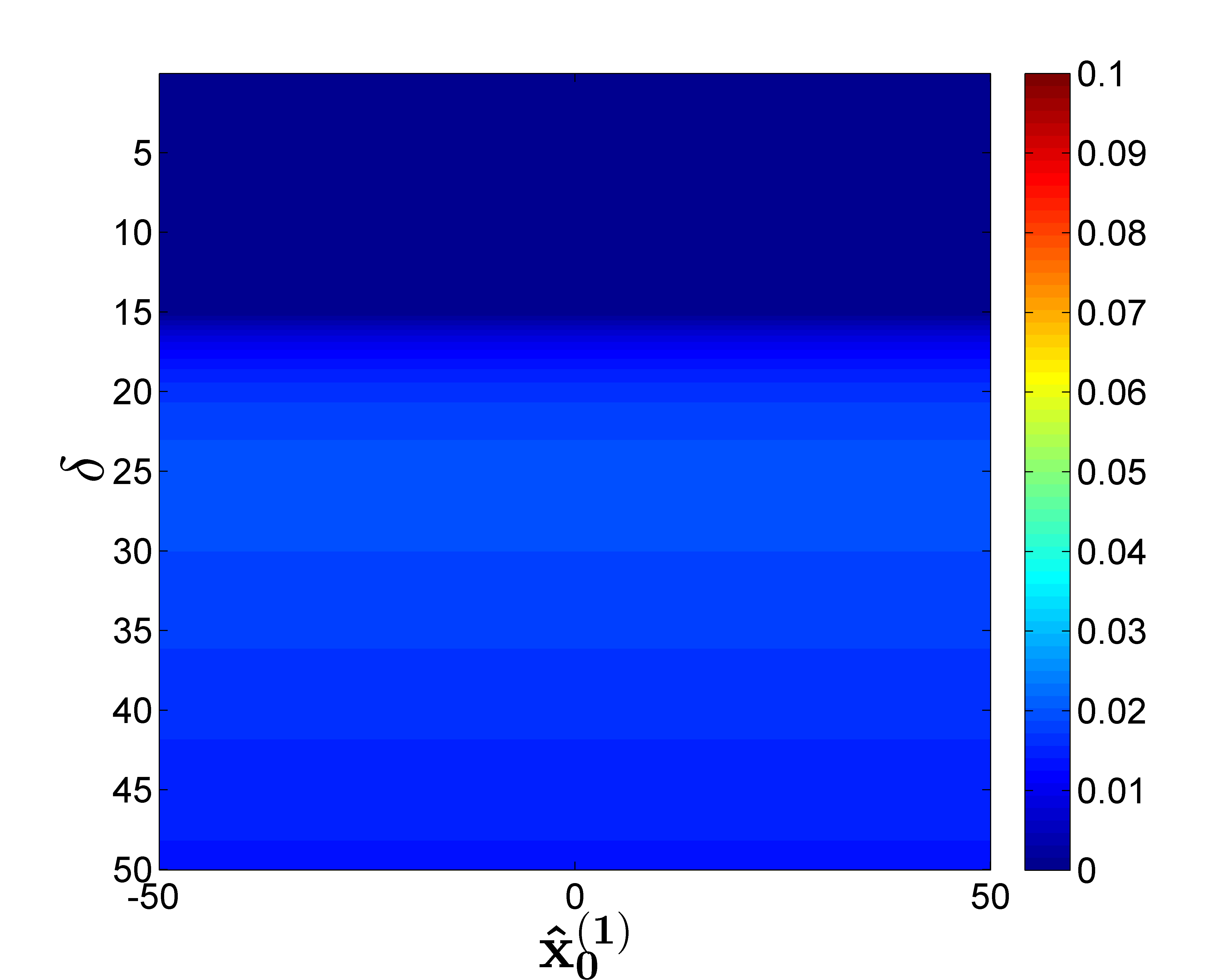} %
			\label{Fig:transform_coding__optimal_linearization__transform_domain_analysis__component_2__a22}}}
	\\	
		{\subfloat[$\transU{b}_{1}^*$]{\includegraphics[width=0.23\textwidth]{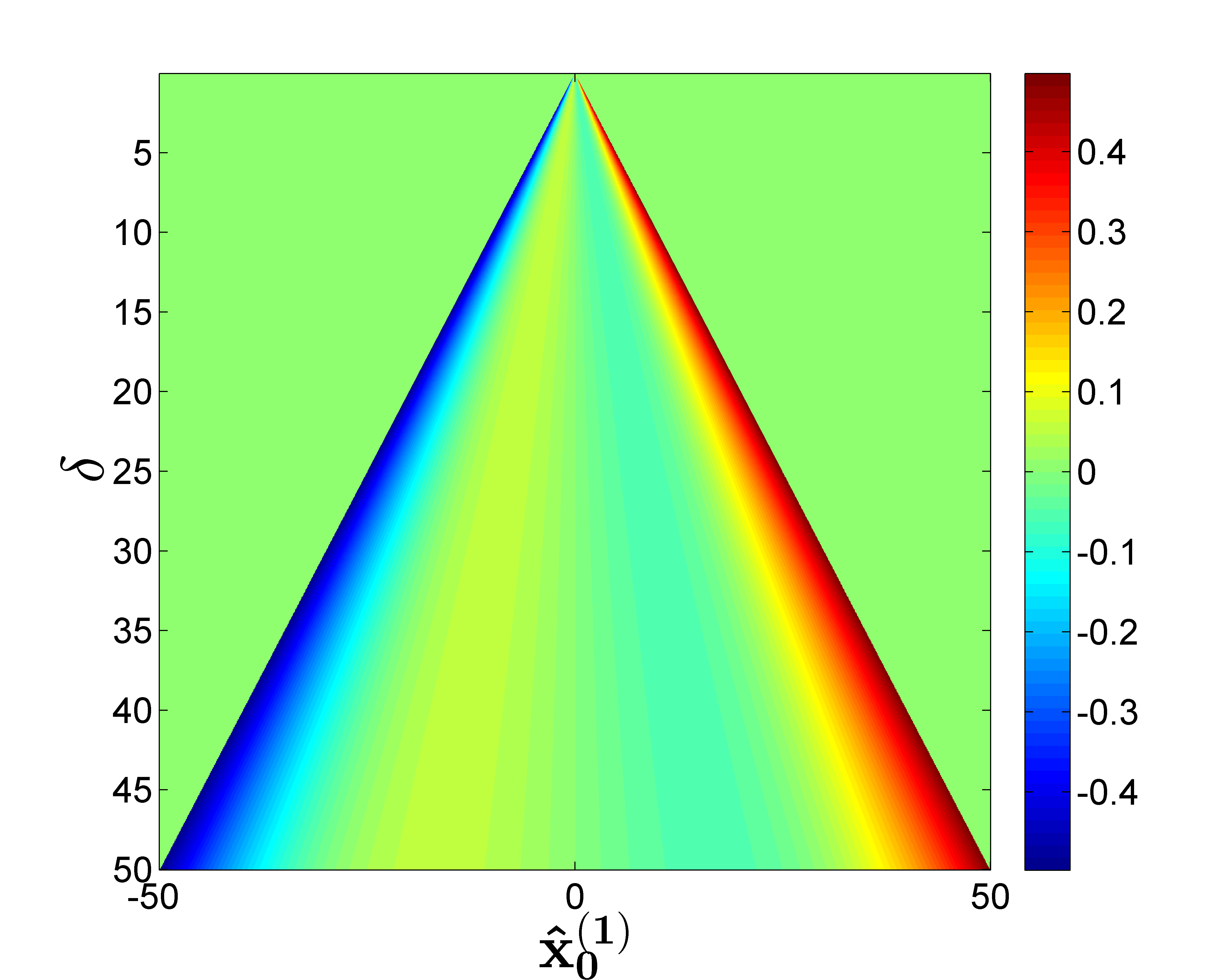} %
				\label{Fig:transform_coding__optimal_linearization__transform_domain_analysis__component_1__b1}} }
		{\subfloat[$\transU{b}_{2}^*$]{\includegraphics[width=0.23\textwidth]{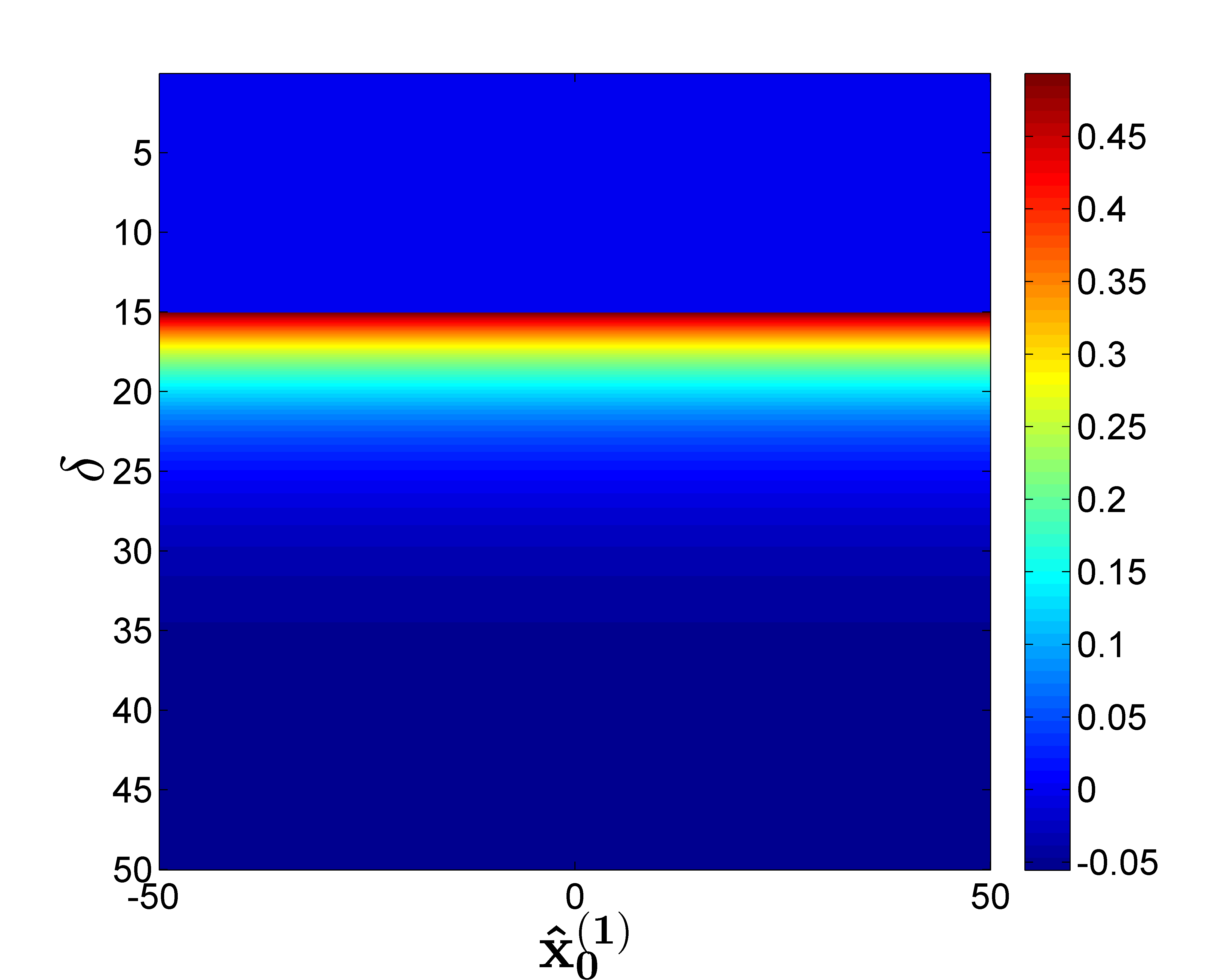} %
				\label{Fig:transform_coding__optimal_linearization__transform_domain_analysis__component_2__b2}}}
		\\
			{\subfloat[${LMSE^{SQ}\left(\transU{a}_{11}^*,\transU{b}_{1}^*; {\eta}(\transU{\vec{x}}_0^{(1)},\delta)\right)}$]{\includegraphics[width=0.23\textwidth]{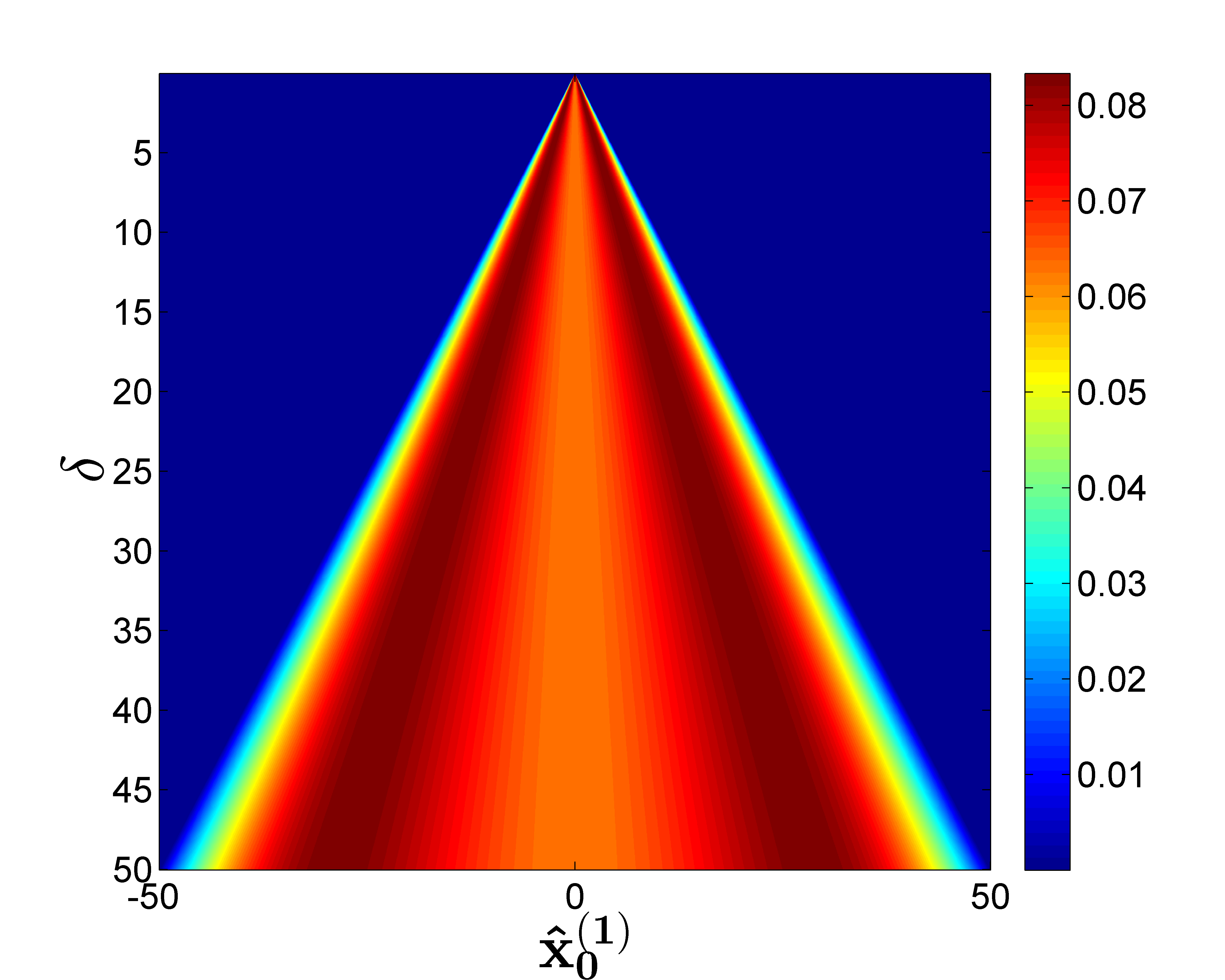} %
					\label{Fig:transform_coding__optimal_linearization__transform_domain_analysis__MSE_of_component_1}} }
			{\subfloat[$~~LMSE^{SQ}\left(\transU{a}_{22}^*,\transU{b}_{2}^*; {\eta}(\transU{\vec{x}}_0^{(2)},\delta)\right)$]{\includegraphics[width=0.23\textwidth]{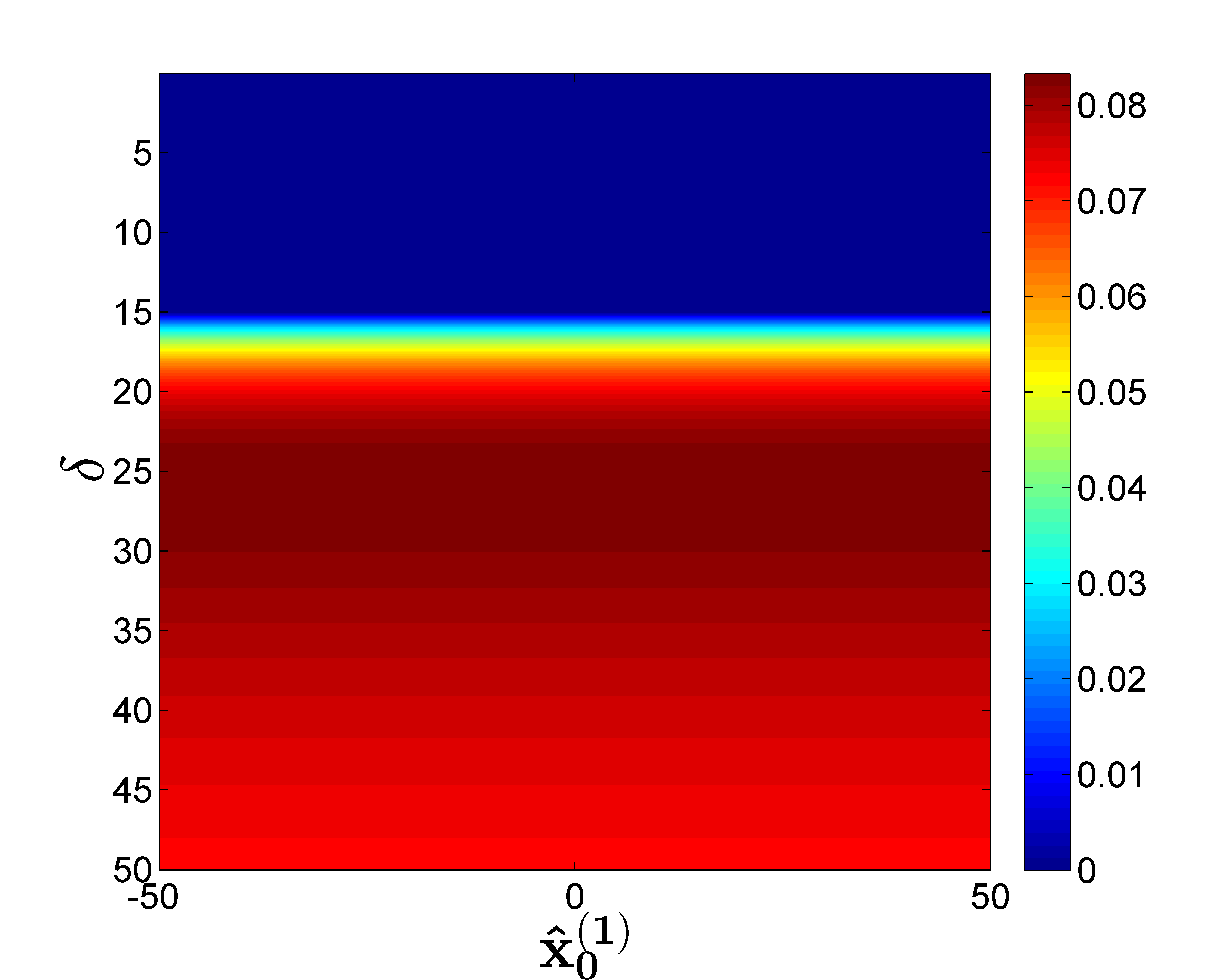} %
					\label{Fig:transform_coding__optimal_linearization__transform_domain_analysis__MSE_of_component_2}}}
			
	\caption{Transform domain parameters of the optimal linear approximation of the exemplary transform coding procedure (for signals in $\mathbb{R} ^2$ and equal transform-domain quantizers). (a)-(b) describe the diagonal elements of the 2x2 matrix $\hat{\mtx{A}}$, and (c)-(d) show the values of $\hat{\vec{b}}$'s components. (e)-(f) show the corresponding approximation errors of the two-transform domain elements.} 
	\label{Fig:Transform domain parameters of the optimal linear approximation of the exemplary transform coding procedure}
\end{figure}

\begin{figure}
	\centering
	{\subfloat[${a}_{11}^*$]{\includegraphics[width=0.23\textwidth]{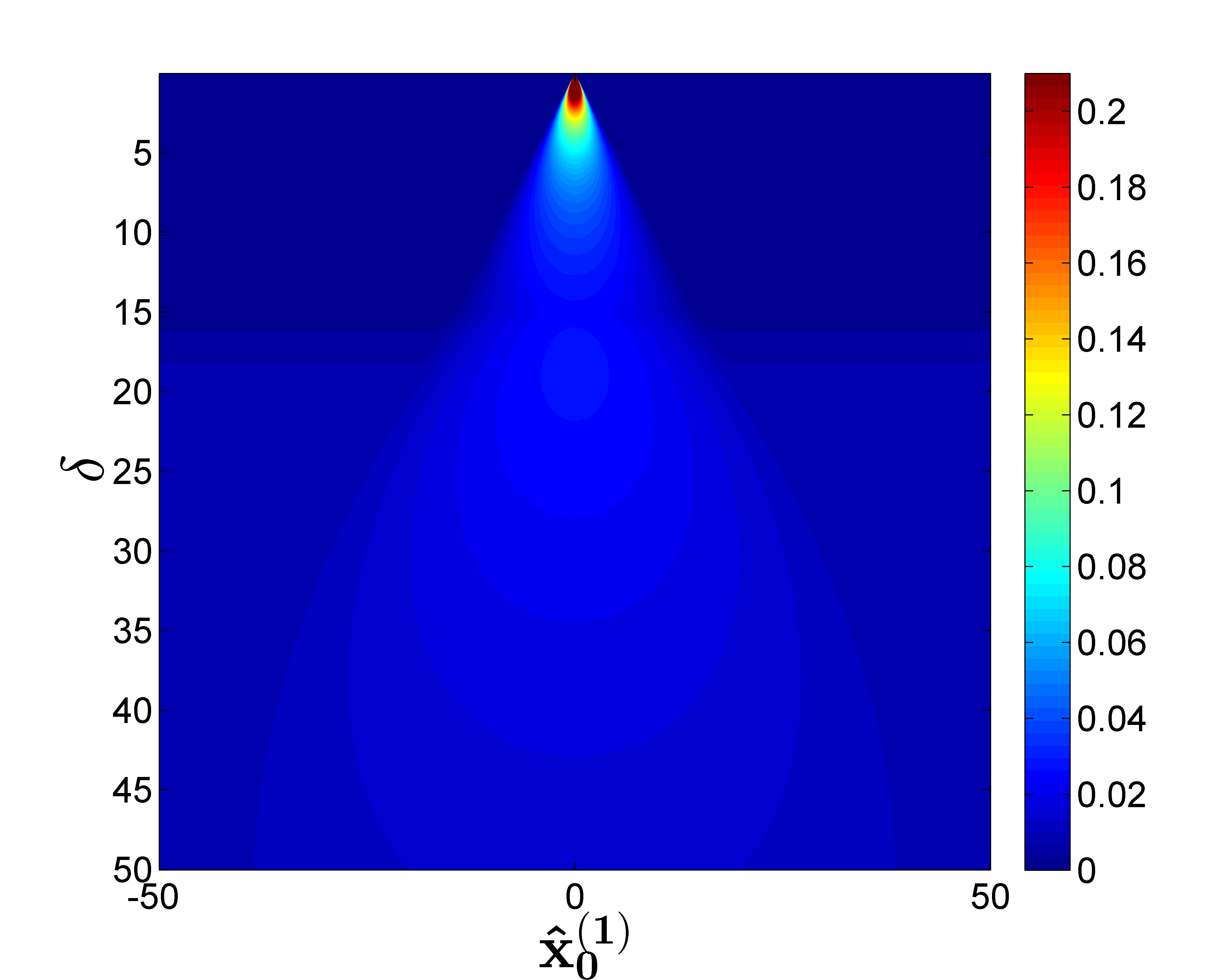} %
			\label{Fig:transform_coding__optimal_linearization__signal_domain_analysis__a11}} }
	{\subfloat[${a}_{12}^*$]{\includegraphics[width=0.23\textwidth]{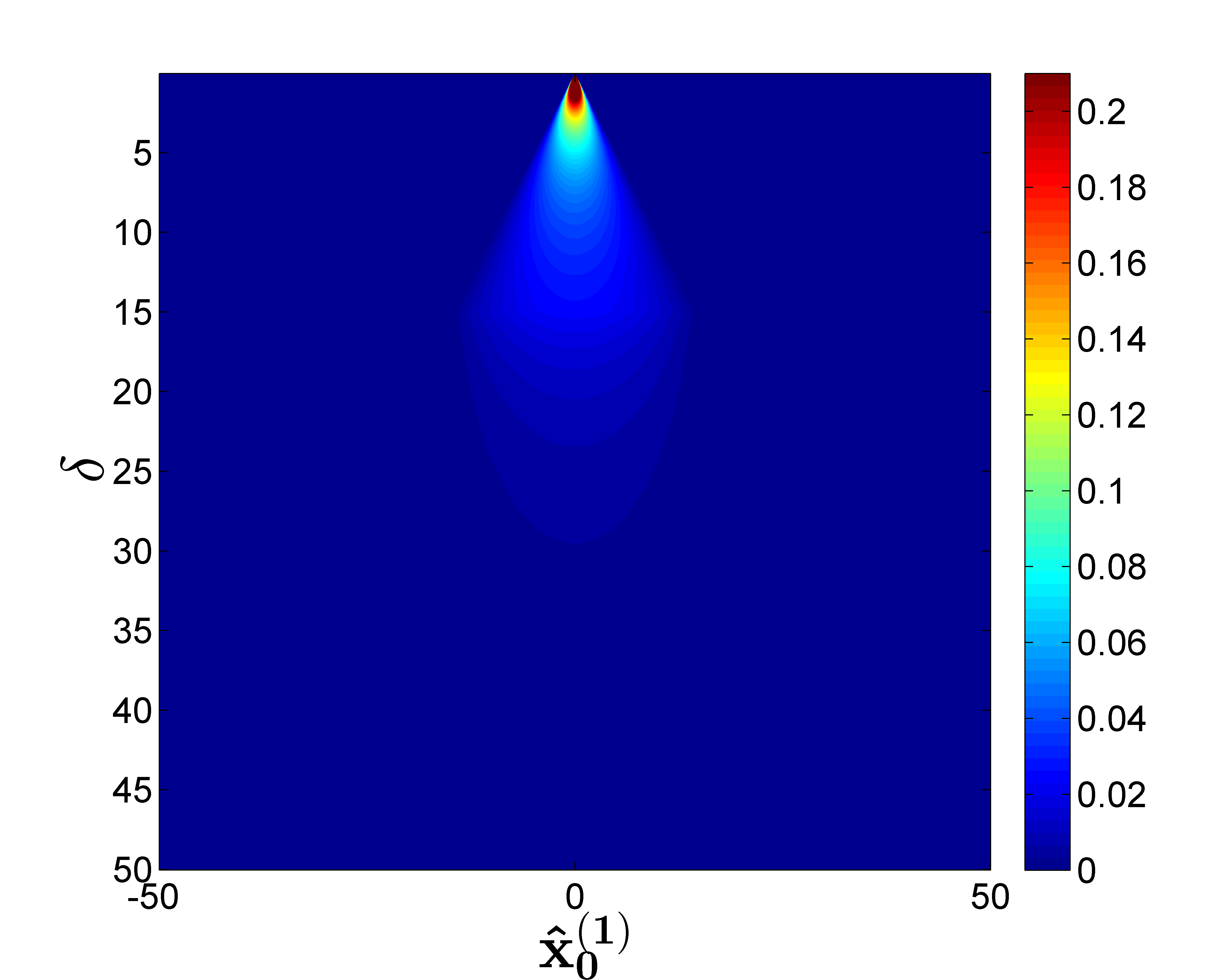} %
			\label{Fig:transform_coding__optimal_linearization__signal_domain_analysis__a12}}}
	\\	
	{\subfloat[${a}_{21}^*$]{\includegraphics[width=0.23\textwidth]{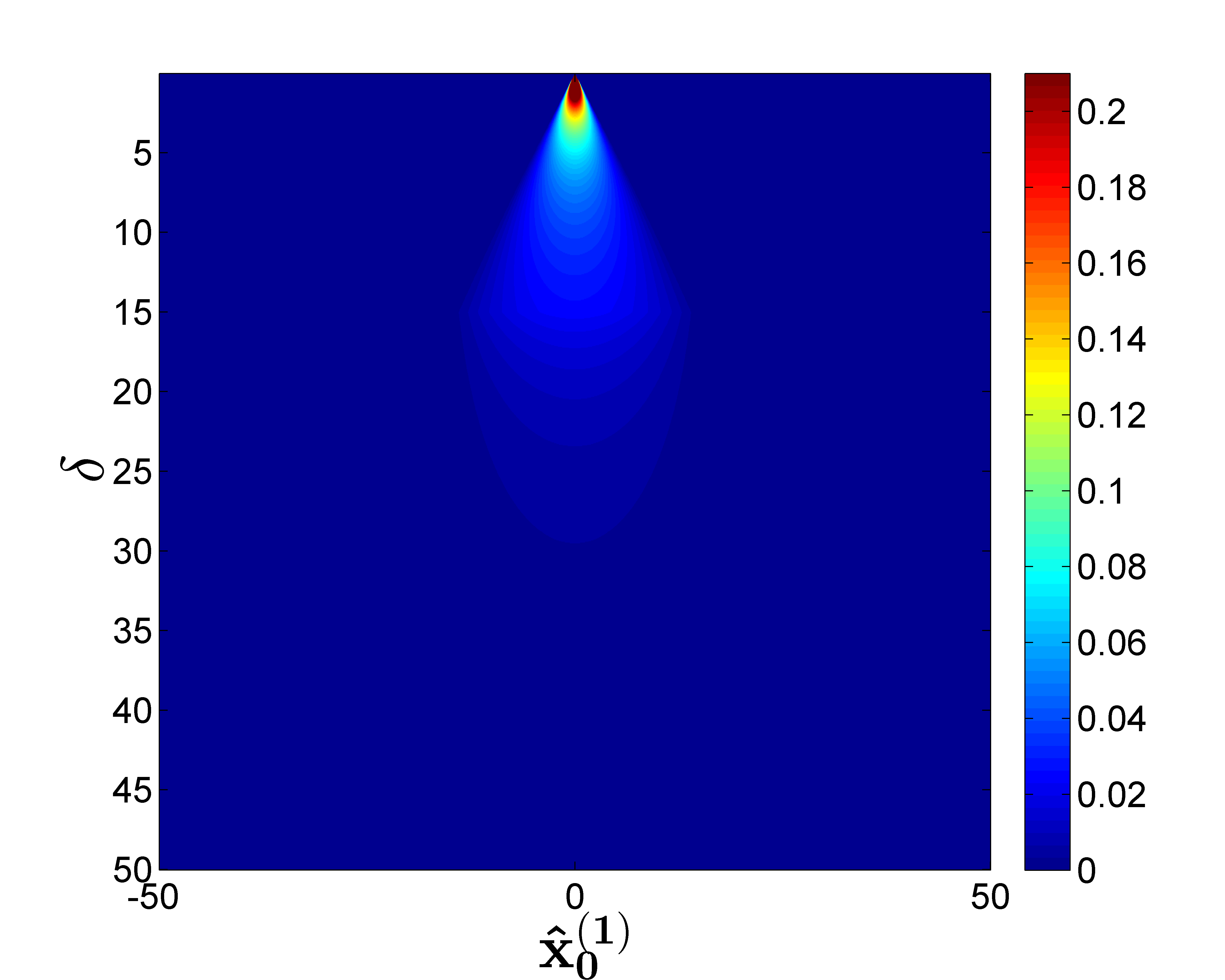} %
			\label{Fig:transform_coding__optimal_linearization__signal_domain_analysis__a21}} }
	{\subfloat[${a}_{22}^*$]{\includegraphics[width=0.23\textwidth]{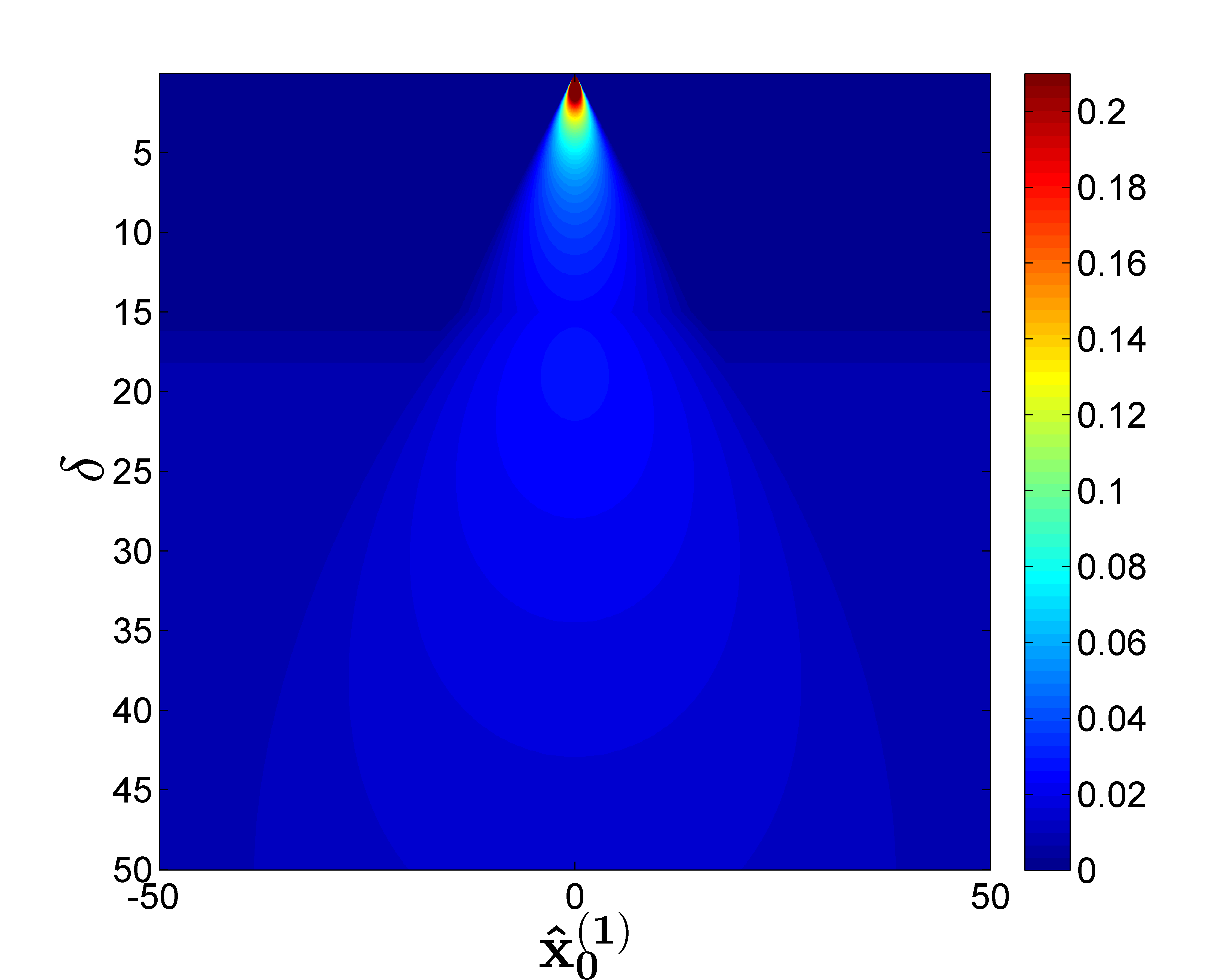} %
			\label{Fig:transform_coding__optimal_linearization__signal_domain_analysis__a22}}}
	\\	
		{\subfloat[${b}_{1}^*$]{\includegraphics[width=0.23\textwidth]{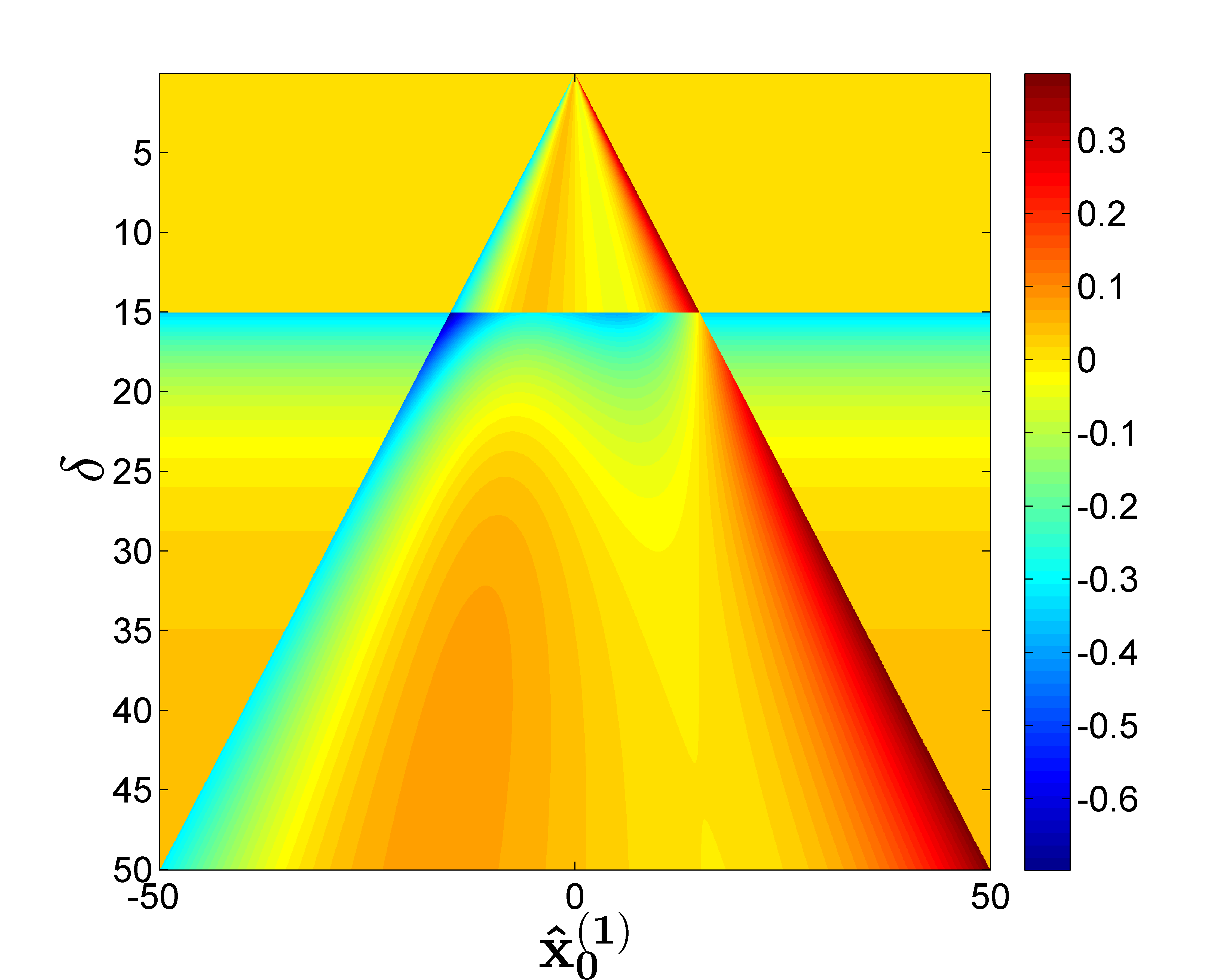} %
			\label{Fig:transform_coding__optimal_linearization__signal_domain_analysis__b1}} }
	{\subfloat[${b}_{2}^*$]{\includegraphics[width=0.23\textwidth]{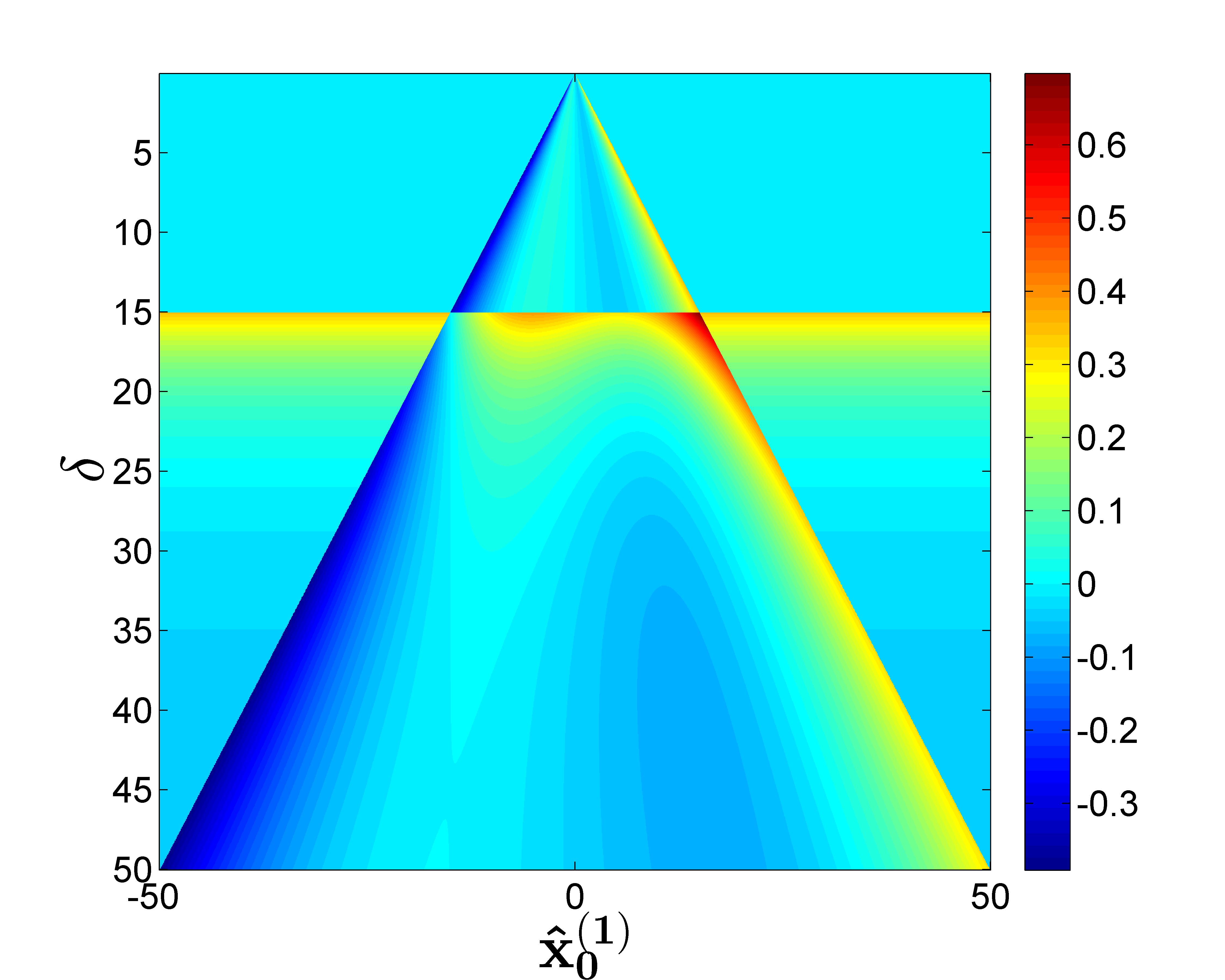} %
			\label{Fig:transform_coding__optimal_linearization__signal_domain_analysis__b2}}}
	\caption{Signal domain parameters of the optimal linear approximation of the exemplary transform coding procedure (for signals in $\mathbb{R} ^2$ and equal transform-domain quantizers). (a)-(d) describe the components of the 2x2 matrix $\mtx{A}$, and (e)-(f) show the values of $\vec{b}$'s components.} 
	\label{Fig:Signal domain parameters of the optimal linear approximation of the exemplary transform coding procedure} 
\end{figure}

Let us exemplify the latter analysis on a transform coder of two-component signals (i.e., $\vec{x}\in\mathbb{R}^2$), that scalary applies the normalized two-level quantizer that was studied above (see Eq. (\ref{eq:Two level quantizer})) on the two components in the domain of the $45^{\circ}$-rotation matrix that takes the 2x2 form of $\mtx{U}_{\pi / 4} = \frac{1}{{\sqrt 2 }}\left[ {\begin{array}{*{20}{c}}
	1&{ - 1} \\ 
	1&1 
	\end{array}} \right]$. The approximation is around $\vec{x}_0 = \mtx{U}_{\pi / 4} {\vec{\transU{x}_0}}$ in a $45^{\circ}$-rotated square neighborhood defined by the  $\eta_U (\vec{x}_0,\delta)$ (see Fig. \ref{Fig:approximation_area_transformation_rotated}). We further define the first component of ${\vec{\transU{x}_0}}$ to vary and fix the second on the value of 15, i.e., ${\vec{\transU{x}_0}} = \left[ {\begin{array}{*{20}{c}}
		{\vec{\transU{x}_0}}^{(1)} \\ 
		{15} 
		\end{array}} \right] $.
The overall linearization error as function of $\delta$ and ${\vec{\transU{x}_0}}^{(1)}$ (Fig. \ref{Fig:Transform coding - 2x2 45degrees rotation - optimal linearization - overall MSE}) shows that it combines the errors of the scalar linearization of the transform coefficients (Figs. \ref{Fig:transform_coding__optimal_linearization__transform_domain_analysis__MSE_of_component_1}-\ref{Fig:transform_coding__optimal_linearization__transform_domain_analysis__MSE_of_component_2}). 
The corresponding parameters in the signal domain (where the matrix $\mtx{A}$ is not necessarily diagonal) are given in Fig. \ref{Fig:Signal domain parameters of the optimal linear approximation of the exemplary transform coding procedure}.
Again, the results generalize the previous observations by demonstrating that minimal MSE is obtained for approximation over the minimal area that includes the nearest non-differentiable point of the compression function (see Fig. \ref{Fig:Transform coding - 2x2 45degrees rotation - optimal linearization - overall MSE}).


\begin{figure}
	\centering
	{\subfloat[$\delta = 0.5$] {\includegraphics[width=0.23\textwidth]{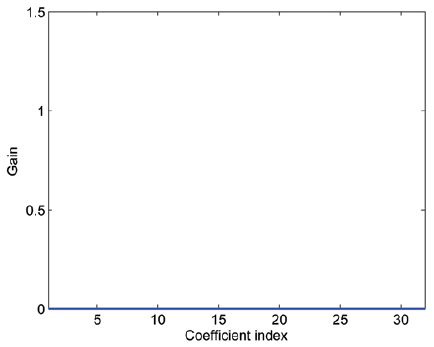} %
			\label{Fig:filter_interpretation_delta_0_5}} }
	{\subfloat[$\delta = 5$] {\includegraphics[width=0.23\textwidth]{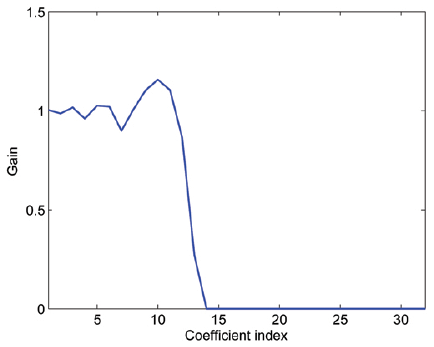} %
			\label{Fig:filter_interpretation_delta_5}} } %
	\\
	{\subfloat[$\delta = 50$] {\includegraphics[width=0.23\textwidth]{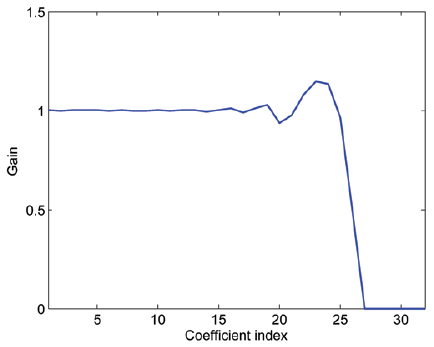} %
			\label{Fig:filter_interpretation_delta_50}} }
	{\subfloat[$\delta = 500$] {\includegraphics[width=0.23\textwidth]{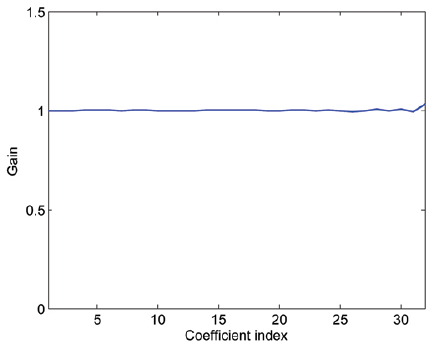} %
			\label{Fig:filter_interpretation_delta_500}} } %
		\caption{Interpretation of a diagonal $\transU{\mtx{A}}^*$ as a transform-domain filter that depends on $\delta$. 
			Here $N=32$ and the $i^{th}$ quantization step is $\Delta_i = 2^{i/4}$.} 
	\label{Fig:filter interpretation - transform domain} 
\end{figure}

We now further develop the discussed transform-coder to the common procedure where the transform-coefficients are uniformly quantized according to different step sizes, $\lbrace \Delta _i \rbrace_{i=1}^{N}$, getting coarser for higher frequencies, i.e., $\Delta_i \le \Delta_j$ for $i < j$. 
Let us consider $N$-length signal vectors and analyze the linear approximation over the $\eta_U (\vec{x}_0,\delta)$ neighborhood, where $\transU{\mtx{A}}^*$ is diagonal. Accordingly, in this case, quantization of each transform-coefficient is linearized separately over a one-dimensional interval of size $2\delta$. However, due to different quantization-steps and approximation-points, the $\lbrace \transU{a}_{ii}^{*}\rbrace_{i=1}^{N}$ values vary. 
We simplify the discussion by approximating around a vector $\vec{x}_0$ with components residing near the middle of the scalar decision-regions of the transform-domain quantizers. Then, relying on the above analysis of the one-dimensional uniform quantizer, we note the following behavior of the sequence $\lbrace \transU{a}_{ii}^{*}\rbrace_{i=1}^{N}$ for $2\delta \approx \Delta_K$ ($K>1$).
First, for some integer $L$ ($1 \le L < K$), the relation $\Delta_{i} \ll  \Delta_{K} \approx 2\delta$ holds for any $i \le L$, and therefore, the optimal parameters are $\transU{a}_{ii}^{*} \approx 1$ and $\transU{b}_{i}^{*} \approx 0$.
Second, for $L < i \le K$, the $2\delta$ value is still greater than $\Delta_i$, however relatively closer, hence, the corresponding $\transU{a}_{ii}^{*}$ values fluctuate.
Finally, the $i > K$ coefficients have quantization steps that are greater than $2\delta$, and accordingly, $\transU{a}_{ii}^{*} \approx 0$.
The latter qualitative analysis lets us to interpret the local-approximation of the transform-coder, $\transU{\mtx{A}}^*$, as a low-pass filter that depends on $\delta$ and the approximation point. Furthermore, the numerical results (Fig. \ref{Fig:filter interpretation - transform domain}) demonstrate the above by showing preservation of low frequencies, an unstable transition phase, and attenuation of high-frequency components.
Note that for too low or too high values of $\delta$ the filter has a all-stop (Fig. \ref{Fig:filter_interpretation_delta_0_5}) or all-pass (Fig. \ref{Fig:filter_interpretation_delta_500}) behavior, respectively.

\begin{figure}
	\includegraphics[width=3.1in]{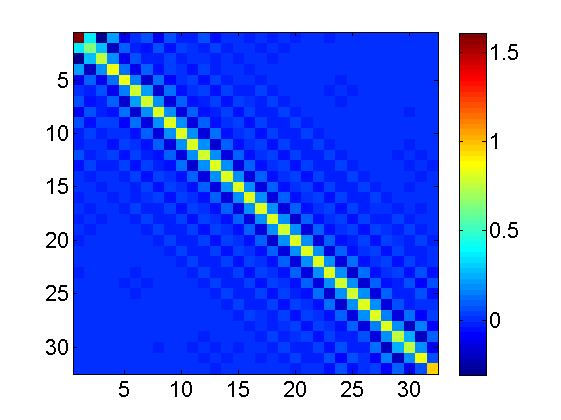}
	\caption{Interpretation of ${\mtx{A}}^*$ as a signal-domain filter that depends on $\delta$. Presented here for the case of $\delta = 50$ from Fig. \ref{Fig:filter_interpretation_delta_50}, incorporated in a DCT-based coding.} 
	\label{Fig:filter interpretation - signal domain - delta 50}
\end{figure}

We conclude by considering the signal-domain filter ${\mtx{A}}^*$ related to $\transU{\mtx{A}}^*$ by the inverse-transformation in (\ref{eq:General transform coding - approximation error - optimality in signal-domain - A}). When the compression utilizes the Discrete Fourier Transform (DFT) and the approximation is over $\eta_U (\vec{x}_0,\delta)$, then the diagonal matrix $\transU{\mtx{A}}^*$ yields a circulant ${\mtx{A}}^*$. 
While the latter involves complex-valued calculations, coding using Discrete Cosine Transform (DCT) keeps the procedure over the reals. The signal-domain filter, ${\mtx{A}}^*$, of the DCT-based coding is exemplified in Fig. \ref{Fig:filter interpretation - signal domain - delta 50} showing an approximately Toeplitz structure.

\section{Experimental Results}
\label{sec:Experimental Results}
In this section we demonstrate the performance of the proposed postprocessing method by presenting results obtained in conjunction with various compression methods. 
We start by considering the simplistic compression procedures of scalar quantization and one-dimensional transform coding. Then, we proceed to the leading image compression standards: JPEG, JPEG2000 and the recent HEVC.

In all experiments we use the BM3D method \cite{RefWorks:169} as the denoiser. 
Since the proposed technique uses a well established denoiser as a subroutine, we compare our method with a single application of this denoiser as a postprocessing procedure. 
This approach is further strengthened by endorsing the denoiser with an oracle capability by searching for the best parameter in terms of maximal PSNR result. More specifically, this oracle denoiser optimizes its output PSNR based on the knowledge of the precompressed image, a capability that cannot be applied in a real postprocessing task.

The computational complexity of our method is mainly determined by the complexity levels of the utilized denoiser and the Jacobian estimation procedure. The latter further depends on the implementation of the compression-decompression method, as it is repeatedly applied according to (\ref{eq:Jacobian estimation - k-th column}). In addition, equation (\ref{eq:Jacobian estimation - k-th column}) exhibits also the effect of the size of the set $S_\delta$ utilized for approximating a single column of the Jacobian.
Since the number of Jacobian columns is as the number of signal samples (denoted as $N$), a straightforward computation of the Jacobian is costly and requires $N$ calculations of (\ref{eq:Jacobian estimation - k-th column}).
Furthermore, the Jacobian matrix is of $N\times N$ size, and is often too large to allow accurate solution of (\ref{eq:iterative solution equations - forward model step - closeness constraint}).
Fortunately, the computational requirements of the estimation of the entire Jacobian matrix can be relaxed for many compression methods that operate independently on adjacent blocks. Specifically, the Jacobian becomes a block-diagonal matrix and, therefore, its columns can be arranged in independent subsets for concurrent computation. This reduces the number of compression-decompression applications to the order of the block size. 
Moreover, the block-diagonal structure of the Jacobian allows to decompose the computation of (\ref{eq:iterative solution equations - forward model step - closeness constraint}) to handle each block separately.
Furthermore, this block-diagonal structure can be assumed even for compression methods that do not conform with it (e.g., JPEG2000), and thus somewhat compromising the postprocessing result, in order to offer a reasonable run-time. 
The (possibly assumed) block size of the compression procedure is denoted here as $B_H \times B_W$, and yields a Jacobian with blocks of size $B_H B_W \times B_H B_W$ along its main diagonal.

The code was implemented in Matlab.
While the settings differ for the various compression methods, a similar stopping criterion is applied. 
In (\ref{eq:method of multipliers form}) we introduced the scaled dual-variable of the $i^{th}$ iteration, ${\vec{u}}_i \in \mathbb{R} ^n$.
We here denote $\Delta \vec{u}_i = \frac{1}{N} \left\| \vec{u}_i - \vec{u}_{i-1} \right\| _1$ and set the algorithm termination conditions to be at one of the following: $\Delta \vec{u}_i < 0.05$, $\Delta \vec{u}_i > \Delta \vec{u}_{i-1}$ or some maximal number of iterations attained.

The remaining parameters are set for each compression method as specified in Table \ref{table:Experimental Results - Parameters}. While the relation between the parameters to the compression method is complex, one can claim that the parameters express the non-differentiable nature of the compression function. For example, HEVC compression, which is an intricate compression method, needs smaller $\delta$ values in the Jacobian approximation and a higher $\mu$, both constraining the linearization to be more local than for the other simpler compression methods.
Furthermore, the parameter settings consider the compression bit-rate, as this quantity reflects the complexity of the given image with respect to the specific compression procedure. Accordingly, the formulas in Table \ref{table:Experimental Results - Parameters} were empirically determined to provide an adequate performance.

\begin{table*} []
	\renewcommand{\arraystretch}{1.3}
	\caption{Experimental settings for the examined compression methods}
	\label{table:Experimental Results - Parameters}
	\centering
	\resizebox{\textwidth}{!}{%
		\begin{tabular}{|c||c|c|c|c|c|c|c|}
			\hline
			\bfseries \shortstack{\\Compression\\Method} & \bfseries \shortstack{\\Affecting\\Factors} & \bfseries \shortstack{\\Max.\\Iterations}& \bfseries $S_{\delta}=\left\lbrace 0.1 \tilde\Delta k \right\rbrace _{k=1}^5$ & \bfseries $\lambda$ & \bfseries $\beta$ & \bfseries $\mu$ & \bfseries $B_H \times B_W$ \\
			\hline\hline
			\shortstack{\\Scalar Quantization\\~} & \shortstack{$r$ bit-rate (bpp)\\$\Delta$ quantizer step size} & 6 & $\tilde\Delta = \Delta $  & 0.01 & $500 \cdot 2^{-2r}$ & $5\cdot 10^{-4} \cdot 2^{0.6 r}$ & $1\times 1$\\
			\hline
			\shortstack{\\Simplistic\\Transform Coding~} &  \shortstack{$\Delta$ quantizer step size (in transform domain)\\effective bit-rate $\tilde r=16-log_2(\Delta)$ } & 10 & $\tilde\Delta = \Delta $  & 0.03 & \shortstack{Aligned approx. area: $200 \cdot 2^{-0.5 \tilde r}$\\Rotated approx. area:$500 \cdot 2^{-0.5 \tilde r}$} & $5\cdot 10^{-5} \cdot 2^{0.8 \tilde r}$ & $2\times 1$\\
			\hline
			\shortstack{\\JPEG\\~} & $r$ bit-rate (bpp) & 8 &  $\tilde\Delta = \nolinebreak 135 /  r $  & 0.15 & $2\cdot r^{-1}$ & $0.01\cdot 2^{r}$ & $8\times 8$ \\
			\hline
			\shortstack{\\JPEG2000\\~} &  $r$ bit-rate (bpp) & 8 &  $\tilde\Delta = \nolinebreak 100 /  r $  & 0.15 & $5\cdot r^{-1}$ & $0.3\cdot 2^{r}$ & $8\times 8$\\
			\hline
			\shortstack{\\HEVC\\~} &  $r$ bit-rate (bpp) & 8 & $\tilde\Delta = \nolinebreak 50 /  r $  & 0.15 & $5\cdot r^{-1}$ & $0.3\cdot 2^{r}$ & $64\times 64$\\
			\hline
		\end{tabular}
	}
\end{table*}

\subsection{Simplistic Compression Procedures}

\subsubsection{Scalar Quantization}
\begin{figure}
	\centering
	{\subfloat[Scalar Quantization (34.64dB)]{\includegraphics[width=0.23\textwidth]{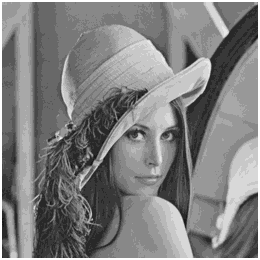}}}
	{\subfloat[Postprocessing Result (37.41dB)]{\includegraphics[width=0.23\textwidth]{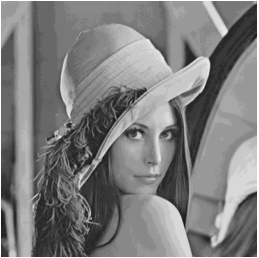}}}
	\caption{Reconstruction of Lena (256x256) from scalar quantization at 4bpp.} 
	\label{Fig:Postprocessing of scalar quantization - Lena 4bpp}
\end{figure}

We begin with the elementary compression procedure of applying uniform scalar quantization (as formulated in (\ref{eq:uniform quantizer formula})) on the signal samples.
Motivated by the analysis in section \ref{subsec:The Case of Multi-Level Uniform Quantization}, we define here the approximation interval to be no longer than the quantization step $\Delta$, as considering only the nearest non-differentiable point yields a useful linearization. Accordingly, the derivative (which is scalar here) is approximated using (\ref{eq:Jacobian estimation - k-th column}) and $S_{\delta}=\left\lbrace 0.1 \Delta k \right\rbrace _{k=1}^5$. 
Our technique achieved impressive PSNR improvements (Table \ref{table:Experimental Results - Scalar Quantization}) over the entire bit-rate range, and consistently passed the oracle denoiser. 
Visually, the false-contouring artifacts were significantly reduced (Fig. \ref{Fig:Postprocessing of scalar quantization - Lena 4bpp}).

\begin{table} []
	\renewcommand{\arraystretch}{1.3}
	\caption{Simplistic Experiment: PSNR Comparison for Scalar Quantization}
	\label{table:Experimental Results - Scalar Quantization}
	\centering
	\begin{tabular}{|c||c|c|c|c|}
		\hline
		\bfseries \shortstack{Image\\256x256} & \bfseries Bit-Rate & \bfseries \shortstack{No\\Postprocessing} & \bfseries \shortstack{Oracle\\Denoiser} & \bfseries \shortstack{Proposed\\Method} \\
		\hline\hline
		\shortstack{Lena} &  2 & 22.86 & 24.54 & \textbf{24.84}  \\
		&  3 & 28.90 & 30.80 & \textbf{31.02} \\
		&  4 & 34.64 & 37.19 & \textbf{37.41}  \\
		&  5 & 40.72 & 42.82 & \textbf{43.02}  \\
		\hline
		\shortstack{Barbara}  &  2 & 23.47 & 25.66 & \textbf{25.93}  \\
		&  3 & 28.48 & 30.94 & \textbf{31.11}  \\
		&  4 & 34.72 & 37.29 & \textbf{37.38}  \\
		&  5 & 40.74 & 42.55 & \textbf{42.64}  \\
		\hline
	\end{tabular}
\end{table}

\subsubsection{One-Dimensional Transform Coding}
\label{subsection: Experimental Results: One-Dimensional Transform Coding}
Now we extend the examined compression method by performing scalar quantization in the transform domain.
Specifically, we split the image into nonoverlapping one-dimensional vertical vectors of 2 pixels. Then we compress them separately by transform coding them using $\mtx{U}_{\pi / 4} = \frac{1}{{\sqrt 2 }}\left[ {\begin{array}{*{20}{c}}
	1&{ - 1} \\ 
	1&1 
	\end{array}} \right]$, followed by applying uniform quantization with identical step size to all the coefficients.
We evaluated here the algorithm performance for the two types of approximation area that were discussed in section \ref{subsec:Linearization analysis for transform coding}: a square area aligned with the axes (i.e., ${\eta}(\vec{x}_0,\delta)$), and a $45^\circ$-rotated squared area (i.e., $\transU{{\eta}}_U (\vec{x}_0,\delta)$) that allows to calculate the linearization in the transform domain and then transforming the parameters back to the signal domain using (\ref{eq:General transform coding - approximation error - optimality in signal-domain - A})-(\ref{eq:General transform coding - approximation error - optimality in signal-domain - b}). Both options used square areas of the same size by setting $S_{\delta}=\left\lbrace 0.1 \Delta k \right\rbrace _{k=1}^5$, where $\Delta$ is the quantizer step size in the transform domain. 
The two area types achieved better results than the oracle denoiser. In that sense, our approach is somewhat robust to the area shape (however, not necessarily to its size). In addition, employing area that is aligned with the axes in the signal domain consistently obtained higher PSNR than the rotated area (Table \ref{table:Experimental Results - Synthetic Transform Coding}). The latter observation will motivate us to use aligned-cubic approximation areas also for more complex compression techniques that will follow next.

\begin{table*} []
	\renewcommand{\arraystretch}{1.3}
	\caption{Simplistic Experiment: PSNR Comparison for Transform Coding of Two-Component Vectors}
	\label{table:Experimental Results - Synthetic Transform Coding}
	\centering
	\begin{tabular}{|c||c|c|c|c|c|}
		\hline
		\bfseries \shortstack{Image\\256x256} & \bfseries Quantizer Step $\Delta$ & \bfseries \shortstack{No\\Postprocessing} & \bfseries \shortstack{Oracle\\Denoiser} & \bfseries \shortstack{Proposed Method\\Aligned Approx. Area} & \bfseries \shortstack{Proposed Method\\Rotated Approx. Area}\\
		\hline\hline
		\shortstack{Lena} &  20  & 31.22 & 35.83 & \textbf{37.00} & 36.78 \\
		&  30  & 27.45 & 32.21  & \textbf{33.37} & 33.00 \\
		&  40  & 24.88 & 29.92  & \textbf{30.92} & 30.29 \\
		\hline
		\shortstack{Barbara} & 15  & 34.50 & 38.00 & \textbf{38.25} & 38.12  \\
		&  20 & 31.79 & 35.65 & \textbf{36.42} & 36.31 \\
		&  30 & 28.05 & 32.33 & \textbf{33.33} & 33.28 \\
		\hline
	\end{tabular}
\end{table*}

\subsection{JPEG}
\label{subsec:Experimental Results - JPEG}
This well known standard \cite{RefWorks:162} is a relatively straight-forward implementation of a two-dimensional transform coding on 8x8 blocks of the image. Specifically, the quantization is performed in the DCT domain where each coefficient has its own quantization step. As the JPEG extends the oversimplified procedure in subsection \ref{subsection: Experimental Results: One-Dimensional Transform Coding}, our postprocessing method is expected to provide good results here. 
Indeed, the experiments show the impressive gains of the suggested method that compete with the prominent techniques from \cite{foi2007pointwise,zhang2013compression} (see Table \ref{table:Experimental Results - JPEG}).
The comparison to \cite{foi2007pointwise,zhang2013compression} indirectly considers additional methods such as \cite{RefWorks:141,RefWorks:144,RefWorks:159,sun2007postprocessing} that were already surpassed by  \cite{foi2007pointwise} and/or \cite{zhang2013compression}.
Moreover, while many competitive methods (e.g., \cite{RefWorks:141,RefWorks:144,RefWorks:159,foi2007pointwise}) are mainly intended to low bit-rate compression, our method handles the entire bit-rate range and excels for medium and high bit-rates (Table \ref{table:Experimental Results - JPEG}).
The thorough evaluation here is based on PSNR values, as well as on the perceptual metric of Structural Similarity (SSIM) \cite{wang2004image}.

\begin{table*} []
	\renewcommand{\arraystretch}{1.3}
	\caption{JPEG: Result Comparison}
	\label{table:Experimental Results - JPEG}
	\centering
	\begin{tabular}{|c||c|c|c|c|c|c|c|c|c|c|c|}
		\hline
		\multirow{2}{*}{\bfseries \shortstack{Image\\512x512}} & \multirow{2}{*}{\bfseries Bit-Rate} & \multicolumn{2}{|c|}{\bfseries \shortstack{JPEG}} &  \multicolumn{2}{|c|}{\bfseries\shortstack{Oracle\\Denoiser}} & \multicolumn{2}{|c|}{\bfseries\shortstack{Foi et al.\\\cite{foi2007pointwise}}} & \multicolumn{2}{|c|}{\bfseries\shortstack{Zhang et al.\\\cite{zhang2013compression}}} & 
		\multicolumn{2}{|c|}{\bfseries \shortstack{Proposed\\Method}} \\
		\cline{3-12}
		&  & PSNR & SSIM & PSNR & SSIM & PSNR & SSIM & PSNR & SSIM & PSNR & SSIM \\
		\hline\hline
		Lena & 0.173 & 27.33 & 0.7367 & 28.77 & 0.7887 & 28.95 & 0.8040 & \textbf{29.07} & \textbf{0.8054} & 28.90 & 0.7965 \\
	   	     & 0.245 & 30.41 & 0.8183 & 31.82 & 0.8555 & 31.84 &  0.8591 & \textbf{31.97} & \textbf{0.8611} & 31.63 & 0.8518 \\
	   	     & 0.363 & 32.96 & 0.8735 & 34.12 & 0.8932 & 34.06 & 0.8927 & 34.24 & 0.8953 & \textbf{34.32} & \textbf{0.8958} \\
	 	     & 0.511 & 34.76 & 0.9036 & 35.64 & 0.9130 & 35.55 & 0.9124 & 35.82 & 0.9155 & \textbf{35.87} & \textbf{0.9160} \\
	 	     & 0.638 & 35.81 & 0.9188 & 36.52 & 0.9238 & 36.44 & 0.9235 & 36.77 & 0.9270 & \textbf{36.81} & \textbf{0.9271} \\
	 	     & 0.807 & 36.86 & 0.9314 & 37.42 & 0.9334 & 37.34 & 0.9332 & 37.73 & \textbf{0.9374} & \textbf{37.79} & 0.9373 \\
	 	     & 1.157 & 38.54 & 0.9476 & 38.91 & 0.9474 & 38.82 & 0.9466 & \textbf{39.25} & \textbf{0.9511} & 39.23 & 0.9500 \\
		\hline
		Barbara & 0.227 & 23.86 & 0.6642 & \textbf{25.40} & 0.7142 & 24.97 & 0.7212 & 25.33 & \textbf{0.7290} & 24.50 & 0.7008 \\
   				& 0.338 & 25.70 & 0.7710 & 27.26 & 0.8038 & 26.54 &  0.8076 & \textbf{27.32} & \textbf{0.8161} & 26.40 & 0.8063 \\
   				& 0.537 & 28.25 & 0.8559 & 30.03 & 0.8761 & 28.91 & 0.8774 & \textbf{30.19} & 0.8859 & 29.67 & \textbf{0.8904} \\
   				& 0.764 & 30.89 & 0.9060 & 32.49 & 0.9175 & 31.42 & 0.9178 & \textbf{32.80} & 0.9259 & 32.59 & \textbf{0.9286} \\
   				& 0.938 & 32.54 & 0.9273 & 33.94 & 0.9338 & 33.02 & 0.9350 & 34.32 & 0.9419 & \textbf{34.33} & \textbf{0.9439} \\
   				& 1.149 & 34.22 & 0.9442 & 35.34 & 0.9464 & 34.64 & 0.9491 & 35.81 & 0.9545 & \textbf{35.94} & \textbf{0.9559} \\
   				& 1.552 & 36.88 & 0.9625 & 37.65 & 0.9619 & 37.22 & 0.9650 & 38.18 & 0.9681 & \textbf{38.31} & \textbf{0.9681} \\
		\hline				
		Boat & 0.187 & 25.56 & 0.6563 & 26.70 & 0.6962 & 26.75 & 0.7013 & \textbf{26.83} & \textbf{0.7048} & 26.74 & 0.7015 \\
			 & 0.291 & 28.13 & 0.7580 & 29.19 & \textbf{0.7883} & 29.14 & 0.7867 & \textbf{29.26} & 0.7921 & 29.03 & \textbf{0.7883} \\
			 & 0.460 & 30.49 & 0.8301 & 31.37 & 0.8471 & 31.29 &  0.8447 & 31.52 & 0.8501 & \textbf{31.61} & \textbf{0.8550}  \\
			 & 0.663 & 32.35 & 0.8691 & 33.14 & 0.8785 & 33.04 & 0.8779 & 33.35 & 0.8823 & \textbf{33.47} & \textbf{0.8854} \\
			 & 0.825 & 33.50 & 0.8880 & 34.16 & 0.8926 & 34.07 & 0.8938 & 34.38 & 0.8975 & \textbf{34.51} & \textbf{0.8992} \\
			 & 1.035 & 34.63 & 0.9042 & 35.19 & 0.9062 & 35.10 & 0.9071 & 35.40 & \textbf{0.9109} & \textbf{35.50} & \textbf{0.9109} \\
			 & 1.497 & 36.43 & 0.9289 & 36.91 & 0.9292 & 36.82 & 0.9311 & \textbf{37.10} & \textbf{0.9338} & 37.07 & 0.9309 \\
		\hline                     
		Bridge  & 0.231 & 23.06 & 0.5713 & 23.76 & 0.5848 & 23.79 & 0.5807 & \textbf{23.85} & \textbf{0.5926} & 23.60 & 0.5921 \\
				& 0.409 & 25.13 & 0.7108 & 25.71 & 0.7175 & 25.72 & 0.7157 & \textbf{25.76} & 0.7234 & 25.62 & \textbf{0.7284} \\
				& 0.688 & 27.01 & 0.8093 & 27.48 & 0.8118 & 27.50 & 0.8157 & \textbf{27.53} & 0.8179 & \textbf{27.53} & \textbf{0.8210} \\
				& 1.003 & 28.50 & 0.8634 & 28.88 & 0.8603 & 28.88 & 0.8692 & 29.00 & 0.8702 & \textbf{29.02} & \textbf{0.8726} \\
				& 1.261 & 29.54 & 0.8917 & 29.95 & 0.8893 & 29.89 & 0.8976 & 30.11 & 0.8989 & 29.92 & \textbf{0.8999} \\
				& 1.579 & 30.80 & 0.9166 & 31.23 & 0.9153 & 31.09 & 0.9215 & \textbf{31.41} & \textbf{0.9233} & 31.17 & \textbf{0.9233} \\
				& 2.197 & 33.32 & 0.9490 & 33.78 & 0.9474 & 33.57 & 0.9522 & \textbf{34.00} & \textbf{0.9544} & 33.72 & 0.9540 \\
		\hline	
		                    
		Peppers & 0.176 & 27.17 & 0.7078 & 28.75 & 0.7748 & 29.04 & \textbf{0.7906} & \textbf{29.17} & 0.7890 & 28.86 & 0.7802 \\
				& 0.246 & 30.14 & 0.7839 & 31.50 & 0.8273 & 31.69 & \textbf{0.8322} & \textbf{31.77} & 0.8316 & 31.68 & 0.8316 \\
				& 0.361 & 32.44 & 0.8354 & 33.45 & 0.8564 & 33.52 & 0.8561 & \textbf{33.67} & \textbf{0.8590} & 33.63 & 0.8584 \\
				& 0.513 & 33.92 & 0.8651 & 34.61 & 0.8730 & 34.64 & 0.8722 & \textbf{34.83} & \textbf{0.8775} & 34.81 & 0.8763 \\
				& 0.650 & 34.77 & 0.8809 & 35.30 & 0.8829 & 35.31 & 0.8824 & \textbf{35.55} & \textbf{0.8892} & \textbf{35.55} & 0.8878 \\
				& 0.832 & 35.59 & 0.8949 & 35.96 & 0.8941 & 35.94 & 0.8927 & \textbf{36.19} & \textbf{0.9001} & 36.17 & 0.8980 \\
				& 1.203 & 36.79 & 0.9135 & 37.03 & 0.9111 & 37.01 & 0.9105 & \textbf{37.23} & \textbf{0.9161} & 37.13 & 0.9130 \\
		\hline				
		\end{tabular}
\end{table*}

\begin{figure}
	\centering
	{\subfloat[JPEG (32.96dB)]{\includegraphics[width=0.23\textwidth]{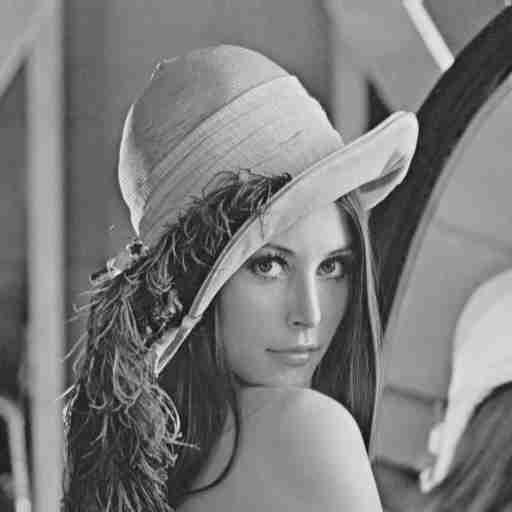}}}
	{\subfloat[Postprocessing Result (34.32dB)]{\includegraphics[width=0.23\textwidth]{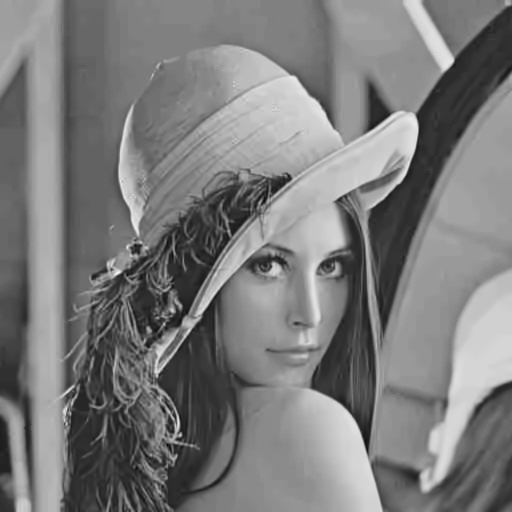}}}
	\caption{Reconstruction of Lena (512x512) from JPEG compression at 0.363bpp.} 
	\label{Fig:Postprocessing of JPEG - Lena 0.271bpp}
\end{figure}

Since JPEG applies transform coding on non-overlapping 8x8 blocks, its Jacobian matrix is indeed block diagonal. 
In addition, the sufficiently small blocks provide a computationally efficient structure that does not need to be simplified further. 
Consequently, the run-time of the JPEG postprocessing (Matlab implementation) is rather reasonable and is usually about 1-2 minutes for a 512x512 image.

\subsection{JPEG2000}
\label{subsec:Experimental Results - JPEG2000}
\begin{figure}
	\centering
	{\subfloat[JPEG2000 (30.79dB)]{\includegraphics[width=0.23\textwidth]{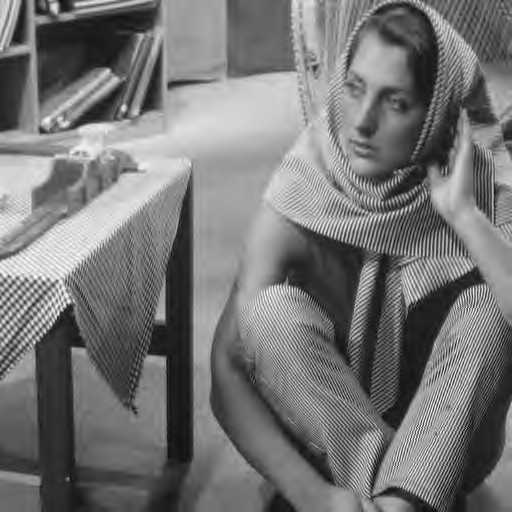}}}
	{\subfloat[Postprocessing Result (31.51dB)]{\includegraphics[width=0.23\textwidth]{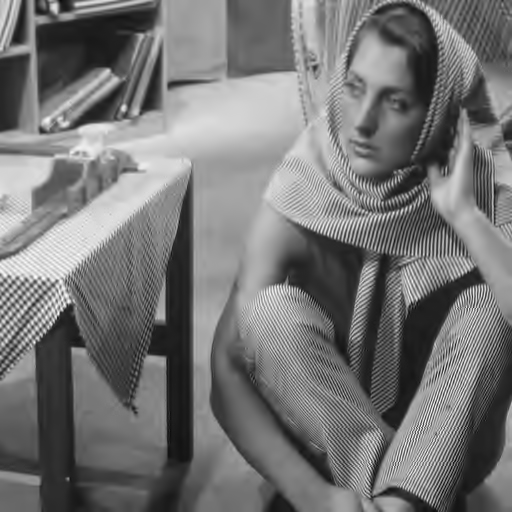}}}
	\\
	{\subfloat[JPEG2000 (Zoomed-in) ]{\includegraphics[width=0.23\textwidth]{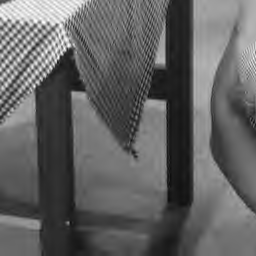}}}
	{\subfloat[Postprocessing Result (Zoomed-in)]{\includegraphics[width=0.23\textwidth]{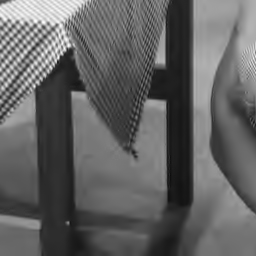}}}
	\caption{Reconstruction of Barbara (512x512) from JPEG2000 compression at 0.40bpp.} 
	\label{Fig:Postprocessing of JPEG2000 - Barbara 0.40bpp}
\end{figure}

This efficient standard \cite{RefWorks:163} applies transform coding in the wavelet domain for relatively large signal blocks (also known as tiles) of at least 128x128 size. Not only the tile size affects the compression run-time, it also impairs the suggested parallelism optimization, as it is beneficial for small block sizes. Nevertheless, it is still recommended to reduce the computational cost by concurrent computation of the Jacobian columns in relatively large subgroups that inevitably contain dependent elements.
Our experiments included postprocessing of images compressed using JPEG2000 compression (via the Kakadu software \cite{jpeg2000_software_Kakadu}) without any tiling. However, the Jacobian was estimated by assuming independent 8x8 blocks, where this reduced accuracy yielded considerable relief in the computational burden (the postprocessing took 5-8 minutes for a 512x512 image).
The reconstruction PSNR of our method reached up to 0.7dB improvement of the JPEG2000 output (e.g., see Fig. \ref{Fig:Postprocessing of JPEG2000 - Barbara 0.40bpp}).
The results in \cite{zhai2009efficient,Kwon2015Efficient} were provided to postprocessing of low bit-rate compression. Therefore, we first compare our results to these from \cite{zhai2009efficient,Kwon2015Efficient} (Table \ref{table:Experimental Results - JPEG2000 - Low Bit-Rates}) according to their experimental settings, and then show results for higher bit-rates where our method is even more effective (Table \ref{table:Experimental Results - JPEG2000 - high bit rates}).
Table \ref{table:Experimental Results - JPEG2000 - Low Bit-Rates} exhibits that our method outperforms \cite{zhai2009efficient} and competitive with the technique from \cite{Kwon2015Efficient}.
In addition, our results for higher bit-rates (Table \ref{table:Experimental Results - JPEG2000 - high bit rates}) compete with the oracle denoiser. These results are encouraging since the oracle denoiser needs the precompressed image and therefore not suitable for the common compression applications.
Furthermore, these results in Tables \ref{table:Experimental Results - JPEG2000 - Low Bit-Rates} and \ref{table:Experimental Results - JPEG2000 - high bit rates} establish our technique as suitable for a wide range of bit-rates.
The restoration results visually demonstrated the artifact reduction using our method, specifically, handling of the ringing artifact (Fig. \ref{Fig:Postprocessing of JPEG2000 - Barbara 0.40bpp}).

\begin{table} []
\begin{threeparttable}
	\caption{JPEG2000: Comparison of PSNR Gains at Low Bit-Rates}
	\label{table:Experimental Results - JPEG2000 - Low Bit-Rates}	
	\centering
		\begin{tabular}{|c||c|c|c|c|c|c|}
			\hline
			{\bfseries \shortstack{Image\\512x512}\tnote{1}} & {\bfseries \shortstack{Bit\\Rate}\tnote{2}} & {\bfseries \shortstack{JPEG2000}\tnote{3}} &   {\bfseries\shortstack{Zhai et al.\\\cite{zhai2009efficient}}} &
			{\bfseries\shortstack{Kwon et al.\\\cite{Kwon2015Efficient}}} & 
			{\bfseries \shortstack{Proposed\\Method}} \\
			\hline\hline
			Lena & 0.10  & 29.86 & 0.03  & \textbf{0.50} & 0.40  \\
			& 0.15  & 31.63 & -0.34  & \textbf{0.54} & 0.39  \\
			& 0.20  & 33.01 & 0.08  & NA & \textbf{0.47}  \\
			\hline
			Peppers & 0.10  & 29.58 & 0.31 & \textbf{0.47} & \textbf{0.47} \\
			& 0.15  & 31.33 & 0.04 & \textbf{0.62} & 0.41  \\
			& 0.20  & 32.51 & -0.27 & NA & \textbf{0.46}  \\
			\hline
			Bridge  & 0.10  & 22.81  & -0.05 & 0.04 & \textbf{0.06}  \\
			& 0.15  & 23.76  & -0.13 & \textbf{0.10} & 0.09  \\
			& 0.20  & 24.34  & -0.17 & NA & \textbf{0.11}  \\
			\hline
		\end{tabular}
		 \begin{tablenotes}
				\item[1] {The results presented in \cite{zhai2009efficient,Kwon2015Efficient} for JPEG2000 postprocessing are for somewhat different versions of the widely-used Lena, Peppers and Bridge images. Accordingly, in this table, and only here, we refer to these images that were specified in \cite{Kwon2015Efficient}.}
				\item[2] {The bit-rate here is the input given to the Kakadu software, and is not necessarily the accurate output bit-rate.}
				\item[3] {As explained in \cite{Kwon2015Efficient}, the PSNR of the compression using the Kakadu software are slightly different in the various papers. Accordingly, the comparison is for the PSNR gains.}
			\end{tablenotes}
  \end{threeparttable}
\end{table}

\begin{table} []
	\renewcommand{\arraystretch}{1.3}
	\caption{JPEG2000: Result Comparison at Medium/High Bit-Rates}
	\label{table:Experimental Results - JPEG2000 - high bit rates}
	\centering
	\resizebox{\columnwidth}{!}{%
	\begin{tabular}{|c||c|c|c|c|c|c|c|}
		\hline
		\multirow{2}{*}{\bfseries \shortstack{Image\\512x512}} & \multirow{2}{*}{\bfseries \shortstack{Bit\\Rate}} & \multicolumn{2}{|c|}{\bfseries \shortstack{JPEG2000}} &  \multicolumn{2}{|c|}{\bfseries\shortstack{Oracle\\Denoiser}} & 
		\multicolumn{2}{|c|}{\bfseries \shortstack{Proposed\\Method}} \\
		\cline{3-8}
		&  & PSNR & SSIM & PSNR & SSIM & PSNR & SSIM \\
		\hline\hline
		                
		Lena & 0.30 & 34.87 & 0.8983 & 35.20 & 0.9005 & \textbf{35.24} & \textbf{0.9019} \\
			& 0.40 & 36.12 & 0.9142 & \textbf{36.43} & 0.9160 & 36.42 & \textbf{0.9161} \\
			& 0.50 & 37.26 & 0.9255 & \textbf{37.49} & \textbf{0.9262} & 37.39 & 0.9254 \\
			& 0.60 & 37.97 & 0.9326 & \textbf{38.16} & \textbf{0.9331} & 37.99 & 0.9313 \\
			& 0.70 & 38.69 & 0.9392 & \textbf{38.87} & \textbf{0.9398} & 38.58 & 0.9370 \\
		\hline		                  		                               
		Barbara & 0.30 & 29.16 & 0.8506 & \textbf{30.00} & 0.8607 & 29.76 & \textbf{0.8635} \\
				& 0.40 & 30.79 & 0.8791 & \textbf{31.70} & 0.8897 & 31.51 & \textbf{0.8915} \\
				& 0.50 & 32.16 & 0.9049 & \textbf{33.09} & 0.9131 & 32.81 & \textbf{0.9134} \\
				& 0.60 & 33.30 & 0.9199 & \textbf{34.20} & 0.9263 & 34.03 & \textbf{0.9277} \\
				& 0.70 & 34.37 & 0.9296 & \textbf{35.17} & 0.9342 & 35.05 & \textbf{0.9358} \\
		\hline		                                               
		Boat & 0.30 & 30.87 & 0.8204 & 31.22 & 0.8237 & \textbf{31.25} & \textbf{0.8272} \\
			& 0.40 & 32.29 & 0.8508 & 32.61 & 0.8535 & \textbf{32.62} & \textbf{0.8557} \\
			& 0.50 & 33.32 & 0.8710 & 33.58 & 0.8729 & \textbf{33.63} & \textbf{0.8750} \\
			& 0.60 & 34.17 & 0.8858 & 34.44 & 0.8864 & \textbf{34.48} & \textbf{0.8889} \\
			& 0.70 & 34.90 & 0.8975 & 35.13 & 0.8973 & \textbf{35.15} & \textbf{0.8993} \\
		\hline		                 
  		 Bridge & 0.30 & 25.42 & \textbf{0.6922} & 25.50 & 0.6848 & \textbf{25.52} & 0.6920 \\
				& 0.40 & 26.36 & \textbf{0.7510} & 26.41 & 0.7397 & \textbf{26.47} & 0.7509 \\
				& 0.50 & 27.24 & 0.7852 & 27.35 & 0.7768 & \textbf{27.39} & \textbf{0.7865} \\
				& 0.60 & 27.89 & 0.8143 & 28.02 & 0.8077 & \textbf{28.04} & \textbf{0.8160} \\
				& 0.70 & 28.50 & 0.8391 & 28.64 & 0.8337 & \textbf{28.65} & \textbf{0.8408} \\
		\hline
		Peppers & 0.30 & 34.11 & 0.8587 & 34.37 & 0.8624 & \textbf{34.39} & \textbf{0.8625} \\
				& 0.40 & 35.05 & 0.8722 & \textbf{35.26} & \textbf{0.8739} & 35.24 & 0.8737 \\
				& 0.50 & 35.80 & 0.8833 & \textbf{35.96} & \textbf{0.8838} & 35.90 & 0.8832 \\
				& 0.60 & 36.34 & 0.8955 & \textbf{36.44} & 0.8936 & 36.42 & \textbf{0.8947} \\
				& 0.70 & 36.83 & 0.9056 & 36.91 & 0.9027 & \textbf{36.93} & \textbf{0.9052} \\
		\hline
		\end{tabular}
	}
\end{table}

\subsection{HEVC}
\label{subsec:Experimental Results - HEVC}
This state-of-the-art coding standard offers a profile of still-image compression \cite{RefWorks:112,RefWorks:164}.
The HEVC applies spatial hybrid-coding on the image by combining a rich prediction capability with transform coding of the prediction residuals. 
In addition, the image is divided into large blocks (also known as coding units) that are further recursively partitioned into rectangular blocks in various sizes. 
Therefore, our Jacobian estimation is set to work on independent blocks of size 64x64, and thus the corresponding run-time was higher than for the previous compression methods. More specifically, postprocessing an HEVC-compressed image took several hours, in contrast to few minutes as needed for the previous compression methods. Accordingly, we stress that our purpose here is to demonstrate the conceptual suitability of our method to compression techniques that are significantly more intricate than transform coding.

\begin{figure}
	\centering
	{\subfloat[HEVC (32.71dB)]{\includegraphics[width=0.23\textwidth]{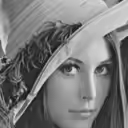}}}
	{\subfloat[Postprocessing Result (32.92dB)]{\includegraphics[width=0.23\textwidth]{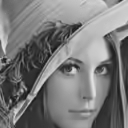}}}
	\caption{Reconstruction of Lena (128x128 Portion) from HEVC compression at 0.639bpp.} 
	\label{Fig:Postprocessing of HEVC - Lena 0.639bpp}
\end{figure}

The results here are for HEVC-compression using the software library in \cite{hevc_software_bpg}.
Again, our postprocessing results reached up to 0.3dB gain in PSNR and often exceeded the oracle denoiser, as shown in the PSNR and SSIM comparison in Table~\ref{table:Experimental Results - HEVC}. 
Figure \ref{Fig:Postprocessing of HEVC - Lena 0.639bpp} visually demonstrates our method's treatment of the delicate artifacts of the HEVC.
To the best of our knowledge, no other artifact-reduction techniques for the HEVC still-image profile have been proposed yet, as it is a recent standard.

\begin{table} []
	\renewcommand{\arraystretch}{1.3}
	\caption{HEVC: Result Comparison}
	\label{table:Experimental Results - HEVC}
	\centering
	\resizebox{\columnwidth}{!}{%
	\begin{tabular}{|c||c|c|c|c|c|c|c|}
		\hline
		\multirow{2}{*}{\bfseries \shortstack{Image\\128x128}} & \multirow{2}{*}{\bfseries \shortstack{Bit\\Rate}} & \multicolumn{2}{|c|}{\bfseries \shortstack{HEVC}} &  \multicolumn{2}{|c|}{\bfseries\shortstack{Oracle\\Denoiser}} & 
		\multicolumn{2}{|c|}{\bfseries \shortstack{Proposed\\Method}} \\
		\cline{3-8}
		&  & PSNR & SSIM & PSNR & SSIM & PSNR & SSIM \\
		\hline\hline		                                       
		Lena & 0.177 & 25.79 & 0.7278 & \textbf{25.91} & 0.7331 & 25.89 & \textbf{0.7337} \\
			& 0.340 & 28.91 & 0.8357 & 28.97 & 0.8349 & \textbf{29.00} & \textbf{0.8359} \\
			& 0.639 & 32.71 & 0.9180 & 32.84 & 0.9181 & \textbf{32.92} & \textbf{0.9202} \\
			& 1.046 & 36.31 & 0.9577 & 36.44 & 0.9572 & \textbf{36.51} & \textbf{0.9583} \\
		\hline		                                               		                               
		Barbara & 0.120 & 27.55 & 0.7658 & 27.71 & 0.7726 & \textbf{27.76} & \textbf{0.7749} \\
				& 0.206 & 30.22 & 0.8420 & 30.39 & \textbf{0.8471} & \textbf{30.45} & 0.8466 \\
				& 0.401 & 33.30 & 0.9163 & 33.48 & 0.9181 & \textbf{33.54} & \textbf{0.9187} \\
				& 0.746 & 36.81 & 0.9554 & 37.04 & 0.9570 & \textbf{37.12} & \textbf{0.9581} \\
		\hline		                                               
		Boat  & 0.212 & 25.29 & 0.7566 & 25.33 & 0.7568 & \textbf{25.35} & \textbf{0.7574} \\
		& 0.416 & 28.54 & 0.8640 & 28.59 & 0.8614 & \textbf{28.67} & \textbf{0.8650} \\
		& 0.735 & 32.44 & 0.9346 & 32.55 & 0.9352 & \textbf{32.68} & \textbf{0.9369}  \\
		& 1.186 & 36.51 & 0.9682 & 36.61 & 0.9684 & \textbf{36.68} & \textbf{0.9692}  \\
		\hline		                                 		
		Bridge  & 0.198 & 25.08 & 0.7172 & \textbf{25.11} & 0.7142 & 25.10 & \textbf{0.7148} \\
				& 0.393 & 28.10 & \textbf{0.8287} & 28.10 & \textbf{0.8287} & \textbf{28.11} & 0.8268 \\
				& 0.746 & 31.64 & \textbf{0.9151} & 31.64 & \textbf{0.9151} & \textbf{31.71} & 0.9144 \\
				& 1.289 & 35.68 & \textbf{0.9653} & 35.69 & \textbf{0.9653} & \textbf{35.71} & 0.9649 \\
		\hline		                        
		Peppers  & 0.1323 & 28.40 & 0.8328 & 28.53 & \textbf{0.8409} & \textbf{28.61} & 0.8408 \\
				& 0.2285 & 31.51 & 0.8947 & 31.71 & \textbf{0.8993} & \textbf{31.81} & 0.8987 \\
				& 0.3813 & 34.73 & 0.9330 & \textbf{34.92} & \textbf{0.9354} & 34.89 & 0.9340 \\
				& 0.6284 & 38.01 & 0.9563 & \textbf{38.18} & \textbf{0.9569} & 38.10 & 0.9560 \\
		\hline
	\end{tabular}
}
	Here the images are the 128x128 portions taken from the center of the 256x256 images.
\end{table}

To summarize this section, the extensive experiments established the proposed compression-artifact reduction technique as a generic method that achieves cutting-edge results for any relevant image compression and over the entire bit-rate range.

\section{Conclusion}
In this paper we proposed a novel postprocessing method for reducing artifacts in compressed images. The task was formulated as a regularized inverse problem, that was subsequently transformed into an iterative form by relying on the ADMM and the Plug-and-Play frameworks. The resulting generic algorithm separately treats the inversion and the regularization, where the latter is implemented by sequentially applying an existing state-of-the-art Gaussian denoiser.
For practicality we simplified the inversion step by representing the nonlinear compression-decompression procedure using a linear approximation. Furthermore, we provided a comprehensive mathematical analysis for linear approximation of simplified quantization and transform-coding operations.
We demonstrated our approach for image compression and presented experimental-results showing impressive gains, that improve upon state-of-the-art postprocessing results for leading image compression standards.


%

%
%

\ifCLASSOPTIONcaptionsoff
  \newpage
\fi



\bibliographystyle{IEEEtran}
\bibliography{IEEEabrv,ADMM_postprocessing_refs}

\begin{thebibliography}{10}
\providecommand{\url}[1]{#1}
\csname url@samestyle\endcsname
\providecommand{\newblock}{\relax}
\providecommand{\bibinfo}[2]{#2}
\providecommand{\BIBentrySTDinterwordspacing}{\spaceskip=0pt\relax}
\providecommand{\BIBentryALTinterwordstretchfactor}{4}
\providecommand{\BIBentryALTinterwordspacing}{\spaceskip=\fontdimen2\font plus
\BIBentryALTinterwordstretchfactor\fontdimen3\font minus
  \fontdimen4\font\relax}
\providecommand{\BIBforeignlanguage}[2]{{%
\expandafter\ifx\csname l@#1\endcsname\relax
\typeout{** WARNING: IEEEtran.bst: No hyphenation pattern has been}%
\typeout{** loaded for the language `#1'. Using the pattern for}%
\typeout{** the default language instead.}%
\else
\language=\csname l@#1\endcsname
\fi
#2}}
\providecommand{\BIBdecl}{\relax}
\BIBdecl

\bibitem{RefWorks:141}
A.~Averbuch, A.~Schclar, and D.~L. Donoho, ``Deblocking of block-transform
  compressed images using weighted sums of symmetrically aligned pixels,''
  \emph{IEEE Trans. Image Process.}, vol.~14, no.~2, pp. 200--212, 2005.

\bibitem{RefWorks:143}
Y.-C. Liaw, W.~Lo, and J.~Z.~C. Lai, ``Image restoration of compressed image
  using classified vector quantization,'' \emph{Pattern Recognition}, vol.~35,
  no.~2, pp. 329--340, 2 2002.

\bibitem{RefWorks:144}
T.~Chen, H.~R. Wu, and B.~Qiu, ``Adaptive postfiltering of transform
  coefficients for the reduction of blocking artifacts,'' \emph{IEEE Trans.
  Circuits Syst. Video Technol.}, vol.~11, no.~5, pp. 594--602, 2001.

\bibitem{RefWorks:146}
K.~Lee, D.~S. Kim, and T.~Kim, ``A new postprocessing algorithm based on
  regression functions,'' in \emph{IEEE International Conference on Acoustics,
  Speech, and Signal Processing (ICASSP)}, vol.~4, 2002, pp.
  IV--3684--IV--3687.

\bibitem{RefWorks:147}
G.~A. Triantaffilidis, D.~Sampson, D.~Tzovaras, and M.~G. Strintzis,
  ``Blockiness reduction in {JPEG} coded images,'' in \emph{International
  Conference on Digital Signal Processing}, vol.~2, 2002, pp. 1325--1328 vol.2.

\bibitem{RefWorks:148}
Y.-C. Liaw, W.~Lo, and J.~Z. Lai, ``Image restoration of compressed image using
  classified vector quantization,'' \emph{Pattern Recognition}, vol.~35, no.~2,
  pp. 329--340, 2002.

\bibitem{RefWorks:150}
F.~Alter, S.~Durand, and J.~Froment, ``Adapted total variation for artifact
  free decompression of {JPEG} images,'' \emph{Journal of Mathematical Imaging
  and Vision}, vol.~23, no.~2, pp. 199--211, 2005.

\bibitem{RefWorks:151}
K.~Du, J.~Lu, H.~Sekiya, Y.~Sun, and T.~Yahagi, ``Post-processing for restoring
  edges and removing artifacts of low bit rates wavelet-based image,'' in
  \emph{International Symposium on Intelligent Signal Processing and
  Communications}, 2006, pp. 943--946.

\bibitem{RefWorks:152}
T.~Kartalov, Z.~A. Ivanovski, L.~Panovski, and L.~J. Karam, ``An adaptive pocs
  algorithm for compression artifacts removal,'' in \emph{International
  Symposium on Signal Processing and Its Applications}, 2007, pp. 1--4.

\bibitem{RefWorks:153}
T.~Tillo and G.~Olmo, ``Data-dependent pre- and postprocessing multiple
  description coding of images,'' \emph{IEEE Trans. Image Process.}, vol.~16,
  no.~5, pp. 1269--1280, 2007.

\bibitem{RefWorks:154}
P.~Weiss, L.~Blanc-Feraud, T.~Andre, and M.~Antonini, ``Compression artifacts
  reduction using variational methods : Algorithms and experimental study,'' in
  \emph{IEEE International Conference on Acoustics, Speech and Signal
  Processing}, 2008, pp. 1173--1176.

\bibitem{RefWorks:155}
K.~Du, H.~Han, and G.~Wang, ``A new algorithm for removing compression
  artifacts of wavelet-based image,'' in \emph{IEEE International Conference on
  Computer Science and Automation Engineering}, vol.~1, 2011.

\bibitem{RefWorks:156}
C.~Jung, L.~Jiao, H.~Qi, and T.~Sun, ``Image deblocking via sparse
  representation,'' \emph{Signal Processing: Image Communication}, vol.~27,
  no.~6, pp. 663--677, 2012.

\bibitem{RefWorks:157}
A.~Zakhor, ``Iterative procedures for reduction of blocking effects in
  transform image coding,'' \emph{IEEE Trans. Circuits Syst. Video Technol.},
  vol.~2, no.~1, pp. 91--95, 1992.

\bibitem{RefWorks:159}
Y.~Yang, N.~P. Galatsanos, and A.~Katsaggelos, ``Regularized reconstruction to
  reduce blocking artifacts of block discrete cosine transform compressed
  images,'' \emph{IEEE Trans. Circuits Syst. Video Technol.}, vol.~3, no.~6,
  pp. 421--432, 1993.

\bibitem{RefWorks:160}
A.~Nosratinia, ``Enhancement of {JPEG}-compressed images by re-application of
  jpeg,'' \emph{Journal of VLSI signal processing systems for signal, image and
  video technology}, vol.~27, no. 1-2, pp. 69--79, 2001.

\bibitem{RefWorks:149}
------, ``Postprocessing of {JPEG}-2000 images to remove compression
  artifacts,'' \emph{IEEE Signal Processing Letters}, vol.~10, no.~10, pp.
  296--299, 2003.

\bibitem{foi2007pointwise}
A.~Foi, V.~Katkovnik, and K.~Egiazarian, ``Pointwise shape-adaptive {DCT} for
  high-quality denoising and deblocking of grayscale and color images,''
  \emph{IEEE Trans. Image Process.}, vol.~16, no.~5, pp. 1395--1411, 2007.

\bibitem{zhang2013compression}
X.~Zhang, R.~Xiong, X.~Fan, S.~Ma, and W.~Gao, ``Compression artifact reduction
  by overlapped-block transform coefficient estimation with block similarity,''
  \emph{IEEE Trans. Image Process.}, vol.~22, no.~12, pp. 4613--4626, 2013.

\bibitem{zhai2009efficient}
G.~Zhai, W.~Lin, J.~Cai, X.~Yang, and W.~Zhang, ``Efficient quadtree based
  block-shift filtering for deblocking and deringing,'' \emph{Journal of Visual
  Communication and Image Representation}, vol.~20, no.~8, pp. 595--607, 2009.

\bibitem{Kwon2015Efficient}
Y.~Kwon, K.~Kim, J.~Tompkin, J.~Kim, and C.~Theobalt, ``Efficient learning of
  image super-resolution and compression artifact removal with semi-local
  gaussian processes,'' \emph{IEEE Trans. on Pattern Analysis and Machine
  Intelligence}, vol.~37, no.~9, pp. 1792--1805, Sept 2015.

\bibitem{RefWorks:138}
M.-Y. Shen and C.~C.~J. Kuo, ``Review of postprocessing techniques for
  compression artifact removal,'' \emph{Journal of Visual Communication and
  Image Representation}, vol.~9, no.~1, pp. 2--14, 3 1998.

\bibitem{RefWorks:167}
M.~V. Afonso, J.~M. Bioucas-Dias, and M.~A.~T. Figueiredo, ``Fast image
  recovery using variable splitting and constrained optimization,'' \emph{IEEE
  Trans. Image Process.}, vol.~19, no.~9, pp. 2345--2356, 2010.

\bibitem{RefWorks:166}
S.~Boyd, N.~Parikh, E.~Chu, B.~Peleato, and J.~Eckstein, ``Distributed
  optimization and statistical learning via the alternating direction method of
  multipliers,'' \emph{Found. Trends Mach. Learn.}, vol.~3, no.~1, pp. 1--122,
  2011.

\bibitem{RefWorks:168}
S.~V. Venkatakrishnan, C.~A. Bouman, and B.~Wohlberg, ``Plug-and-play priors
  for model based reconstruction,'' in \emph{IEEE Global Conference on Signal
  and Information Processing (GlobalSIP)}, 2013, pp. 945--948.

\bibitem{aharon2006KSVD}
M.~Aharon, M.~Elad, and A.~Bruckstein, ``{K-SVD}: An algorithm for designing
  overcomplete dictionaries for sparse representation,'' \emph{IEEE Trans.
  Signal Process.}, vol.~54, no.~11, pp. 4311--4322, 2006.

\bibitem{RefWorks:170}
M.~Elad and M.~Aharon, ``Image denoising via sparse and redundant
  representations over learned dictionaries,'' \emph{IEEE Trans. Image
  Process.}, vol.~15, no.~12, pp. 3736--3745, 2006.

\bibitem{RefWorks:169}
K.~Dabov, A.~Foi, V.~Katkovnik, and K.~Egiazarian, ``Image denoising by sparse
  3-{D} transform-domain collaborative filtering,'' \emph{IEEE Trans. Image
  Process.}, vol.~16, no.~8, pp. 2080--2095, 2007.

\bibitem{RefWorks:162}
G.~K. Wallace, ``The {JPEG} still picture compression standard,'' \emph{IEEE
  Trans. Consum. Electron.}, vol.~38, no.~1, pp. xviii--xxxiv, 1992.

\bibitem{RefWorks:163}
C.~Christopoulos, A.~Skodras, and T.~Ebrahimi, ``The {JPEG}2000 still image
  coding system: an overview,'' \emph{IEEE Trans. Consum. Electron.}, vol.~46,
  no.~4, pp. 1103--1127, 2000.

\bibitem{RefWorks:112}
G.~J. Sullivan, J.~Ohm, W.-J. Han, and T.~Wiegand, ``Overview of the high
  efficiency video coding {(HEVC)} standard,'' \emph{IEEE Trans. on Circuits
  Syst. Video Technol.}, vol.~22, no.~12, pp. 1649--1668, 2012.

\bibitem{RefWorks:164}
J.~Lainema, F.~Bossen, W.-J. Han, J.~Min, and K.~Ugur, ``Intra coding of the
  {HEVC} standard,'' \emph{IEEE Trans. Circuits Syst. Video Technol.}, vol.~22,
  no.~12, pp. 1792--1801, 2012.

\bibitem{RefWorks:165}
T.~Nguyen and D.~Marpe, ``Performance analysis of {HEVC}-based intra coding for
  still image compression,'' in \emph{Picture Coding Symposium}, 2012, pp.
  233--236.

\bibitem{rudin1992nonlinear}
L.~I. Rudin, S.~Osher, and E.~Fatemi, ``Nonlinear total variation based noise
  removal algorithms,'' \emph{Physica D: Nonlinear Phenomena}, vol.~60, no.~1,
  pp. 259--268, 1992.

\bibitem{Schmidhuber15Overview}
J.~Schmidhuber, ``Deep learning in neural networks: An overview,'' \emph{Neural
  Networks}, vol.~61, pp. 85--117, 2015.

\bibitem{gersho2012vector}
A.~Gersho and R.~M. Gray, \emph{Vector quantization and signal
  compression}.\hskip 1em plus 0.5em minus 0.4em\relax Kluwer Academic
  Publishers, 1992.

\bibitem{schuchman1964dither}
L.~Schuchman, ``Dither signals and their effect on quantization noise,''
  \emph{IEEE Transactions on Communication Technology}, vol.~12, no.~4, pp.
  162--165, 1964.

\bibitem{sun2007postprocessing}
D.~Sun and W.-K. Cham, ``Postprocessing of low bit-rate block dct coded images
  based on a fields of experts prior,'' \emph{IEEE Trans. on Image Process.},
  vol.~16, no.~11, pp. 2743--2751, 2007.

\bibitem{wang2004image}
Z.~Wang, A.~C. Bovik, H.~R. Sheikh, and E.~P. Simoncelli, ``Image quality
  assessment: from error visibility to structural similarity,'' \emph{IEEE
  Trans. on Image Process.}, vol.~13, no.~4, pp. 600--612, 2004.

\bibitem{jpeg2000_software_Kakadu}
\BIBentryALTinterwordspacing
``Kakadu software 7.7.'' [Online]. Available:
  \url{http://www.kakadusoftware.com}
\BIBentrySTDinterwordspacing

\bibitem{hevc_software_bpg}
\BIBentryALTinterwordspacing
F.~Bellard, ``{BPG} 0.9.6.'' [Online]. Available: \url{http://bellard.org/bpg/}
\BIBentrySTDinterwordspacing

\end{thebibliography}
%

%
%

%





\end{document}